%% file: report.tex
\documentclass[]{techreport}

\input{config.tex}
\input{preamble.tex}


\hypersetup{
  pdftitle={\reportpdftitle},
  pdfauthor={\reportpdfauthor}
}

\begin{document}

\maketitle
\thispagestyle{firstpage}
\begin{figure}[H]
  \centering
  \includegraphics[width=0.98\linewidth]{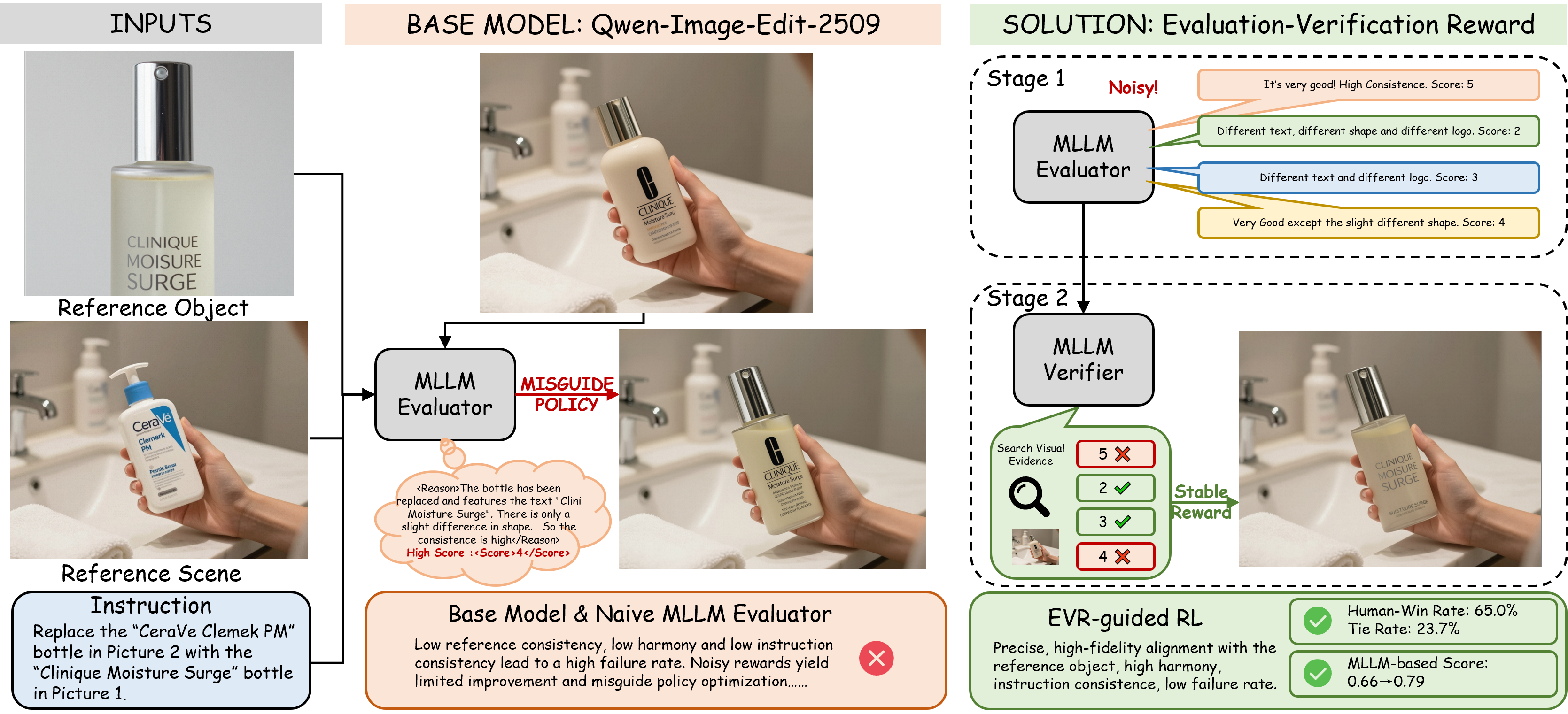}
  \caption{Our EVR framework corrects base model hallucinations via two-stage MLLM verification, scoring then visual grounding, enabling stable reward signals for diffusion-RL. Results show improved alignment, consistency, and human preference over baseline.}
  \label{fig:teaser}
\end{figure}

\input{sections/01_introduction}
\input{sections/02_related_work}
\input{sections/03_method}
\input{sections/05_experiments}

\input{sections/07_conclusion}

\bibliographystyle{plainnat}
\bibliography{references}

\appendix
\thispagestyle{fancy}
\input{sections/appendix/a_implementation_details}
\input{sections/appendix/b_configuration}
\input{sections/appendix/c_release_checklist}
\input{sections/appendix/d_additional_qualitative_results}

\end{document}

%% file: config.tex
\renewcommand{\reportshorttitle}{EVR for Multi-Reference Image Editing}
\renewcommand{\reportdate}{July 2026}
\newcommand{\reportpdftitle}{Evaluation-Verification Reward for Consistent Multi-Reference Image Editing}
\newcommand{\reportpdfauthor}{Yingmao Miao, Pengfei Zhang, Xiaochen Lv, Meng Yu, Lei Sun, Xiangxiang Chu, Chao Shen, and Chenhao Lin}

\setreportlogo{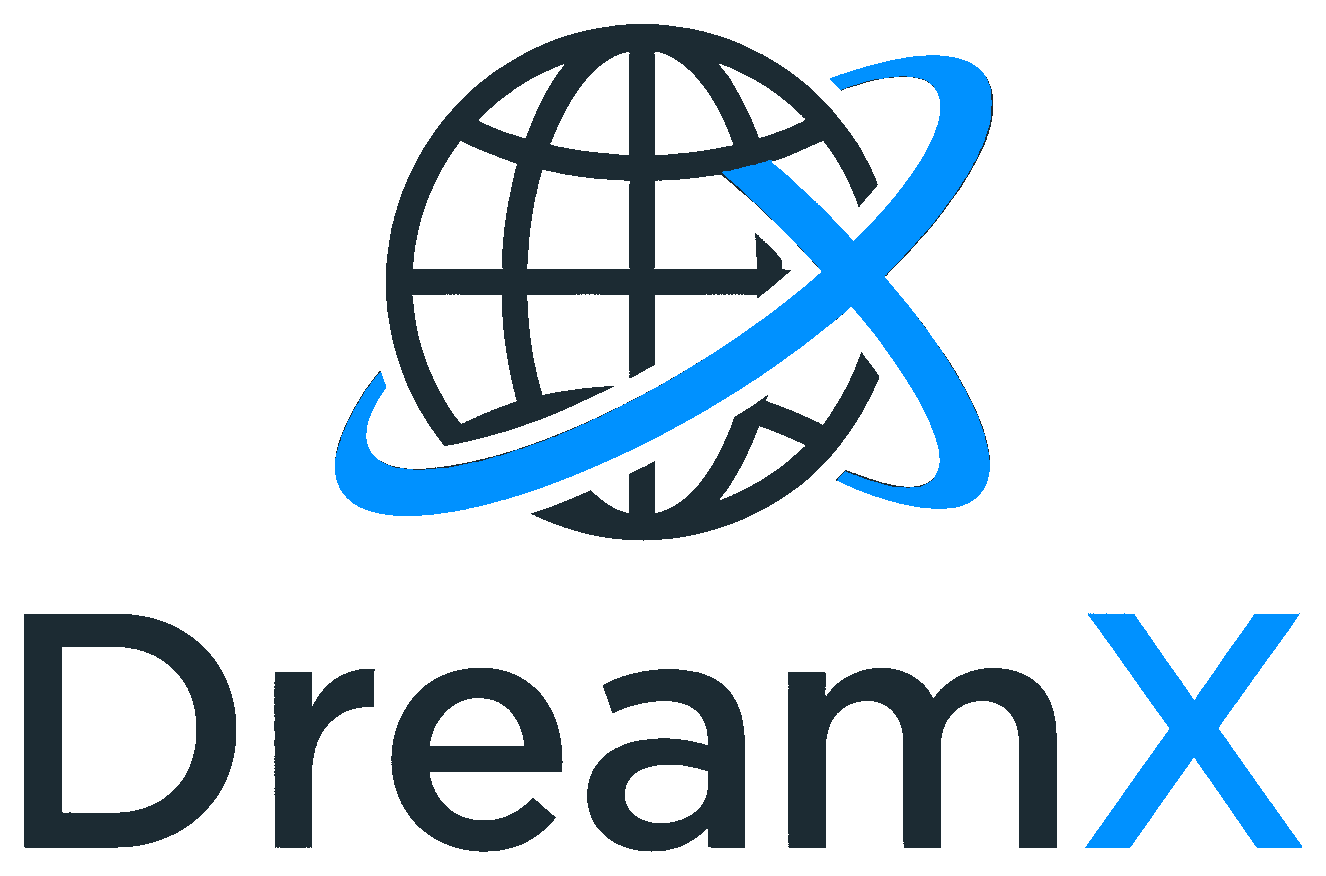}

\title{Evaluation-Verification Reward for Consistent Multi-Reference Image Editing}

\author[1]{Yingmao Miao}
\author[2]{Pengfei Zhang}
\author[1]{Xiaochen Lv}
\author[3]{Meng Yu}
\author[2]{Lei Sun}
\author[2]{Xiangxiang Chu}
\author[1]{Chao Shen}
\author[1]{Chenhao Lin}
\affiliation[1]{Xi'an Jiaotong University}
\affiliation[2]{DreamX Team, Alibaba Group}
\affiliation[3]{Shanghai Jiaotong University}

\abstract{\input{sections/00_abstract}}

\date{\reportdate}

%% file: sections/00_abstract.tex
While recent image editing models have made rapid progress, multi-reference editing remains challenging, particularly in maintaining visual consistency across references and ensuring overall visual harmony. Reinforcement learning (RL) has proven highly effective for text-to-image generation and single-image editing, but its extension to multi-reference editing is hindered by the absence of suitable reward models that capture multi-image relational constraints. Moreover, naively using multimodal large language models (MLLMs) as zero-shot evaluators faces a key tension between hallucination-prone long-form reasoning and the limited deductive power of short-form judgments. We address these issues with a Multi-dimensional Evaluation-Verification Reward (EVR). EVR decomposes evaluation into distinct visual criteria; for each criterion, an MLLM Evaluator generates multiple candidate hypotheses, and a Verifier grounds each claim in concrete visual evidence to accept or reject it, producing reliable and fine-grained reward signals. Together with a scalable data pipeline, our method enables RL fine-tuning of off-the-shelf editors without architectural changes. Extensive experiments show substantial gains over the base Qwen-Image-Edit, improving consistency and harmony to match or surpass NanoBanana.

%% file: preamble.tex
\usepackage{amsmath}
\usepackage{amssymb}
\usepackage{array}
\usepackage{bm}
\usepackage{colortbl}
\usepackage{enumitem}
\usepackage{fancyhdr}
\usepackage{float}
\usepackage{wrapfig}
\usepackage{xurl}
\definecolor{BrandPurple}{HTML}{7366CC}
\definecolor{BrandCyan}{HTML}{00B4E5}
\definecolor{LightGray}{HTML}{F5F5F5}
\definecolor{TableRowAlt}{HTML}{EAF4FB}
\definecolor{HeaderGray}{HTML}{888888}
\definecolor{ForestGreen}{HTML}{228B22}
\definecolor{Gray}{HTML}{808080}

\newcolumntype{L}[1]{>{\raggedright\arraybackslash}m{#1}}
\newcolumntype{C}[1]{>{\centering\arraybackslash}m{#1}}

\newtcolorbox{highlight}{
    colback=BrandCyan!5,
    colframe=BrandCyan,
    arc=2pt,
    boxrule=0.8pt,
    left=6pt, right=6pt, top=4pt, bottom=4pt,
    breakable
}

\newtcolorbox{keyfinding}{
    colback=BrandPurple!5,
    colframe=BrandPurple,
    coltitle=white,
    fonttitle=\bfseries,
    title=Key Finding,
    arc=2pt,
    boxrule=1pt,
    left=6pt, right=6pt, top=4pt, bottom=4pt,
    breakable
}

\newtcolorbox{tipbox}[1]{
    colback=ReportSurface,
    colframe=ReportPrimary,
    coltitle=white,
    fonttitle=\bfseries,
    title=#1,
    arc=2pt,
    boxrule=1pt,
    left=6pt, right=6pt, top=4pt, bottom=4pt,
    breakable
}
\newtcolorbox{promptbox}[1][]{
    enhanced,
    breakable,                  
    colback=gray!10,       
    arc=3pt,                    
    boxrule=0.5pt,              
    left=8pt, right=8pt, top=8pt, bottom=8pt, 
    fontupper=\small,  
    coltitle=black,             
    colbacktitle=gray!10,       
    attach boxed title to top left={yshift=-2mm, xshift=5mm}, 
    boxed title style={boxrule=0.5pt, colframe=black, arc=2pt}, 
}

\fancypagestyle{plain}{%
    \fancyhf{}
    \fancyhead[L]{\footnotesize\color{HeaderGray}\sffamily \reportshorttitle}
    \fancyhead[R]{\footnotesize\color{HeaderGray} \thepage}
    \fancyfoot[C]{}

}

\fancypagestyle{firstpage}{%
    \fancyhf{}
    \fancyhead[L]{\small\sffamily\color{ReportPrimary}
      DreamX Team}
    \fancyhead[R]{\small\color{HeaderGray} \reportdate}
    \fancyfoot[C]{\footnotesize\color{gray}\thepage}

}

%% file: sections/01_introduction.tex
\section{Introduction}
\label{sec:intro}

In recent years, the field of image editing~\citep{controlnet,ipadapter,textualinversion} has witnessed a paradigm shift toward unified models capable of handling diverse tasks within a single framework~\citep{step1x,seedream4,seedream3}. Both open-source methods like Qwen-Image-Edit~\citep{Qwen-Image} and closed-source pioneers such as NanoBanana~\citep{Nanobanana} have demonstrated remarkable versatility in following natural language instructions. However, these models still struggle significantly with multi-reference image editing, a complex setting requiring the synthesis of information from multiple reference images into a harmonious output~\citep{mico}. While Reinforcement Learning (RL) frameworks have proven effective in aligning diffusion models with human preferences for text-to-image generation and single-image editing~\citep{diffusion-dpo,diffusion-kto,flow-grpo,dancegrpo,mixgrpo,tempflowGRPO,diffusionnft,editscore,uniworld}, their potential in multi-reference editing remains largely underexplored. Crucially, the bottleneck in this domain stems from the design of the reward model~\citep{editscore}.

Evaluating multi-reference image editing outputs is inherently challenging. A faithful assessment must be multi-dimensional, covering complex attributes, most notably reference consistency and visual harmony, beyond coarse text alignment and generic visual quality. Yet existing reward models do not meet these requirements. Conventional metrics such as CLIPScore~\citep{clipscore,clip} rely on global semantic embeddings and lack the fine-grained sensitivity needed to judge specific editing attributes. More critically, specialized editing rewards (e.g., EditScore~\citep{editscore,unifiedreward,VQAScore,rewarddance}) are ill-suited to the multi-reference setting: they are trained on human-preference data limited to single-image edits, and thus largely evaluate prompt adherence and overall quality. As a result, they do not generalize to the multi-image relational constraints central to multi-reference editing, including cross-reference consistency and compositional harmony, which are essential for rigorous evaluation.

To close this gap and support multi-dimensional evaluation, recent work increasingly relies on Multimodal Large Language Models (MLLMs)~\citep{llava,qwen2-vl,qwen3vl} as zero-shot evaluators. However, naively applying MLLMs exposes a fundamental trade-off between reasoning depth and visual grounding. As observed in Uniworld-v2(Edit-r1)~\citep{uniworld}, overly long chains-of-thought (CoT) encourage models to lean on internally generated text, resulting in hallucinations rather than evidence-based visual judgments. The issue is further exacerbated when multiple dimensions are evaluated jointly: the expanded CoT budget tends to dilute visual constraints and collapses to trivially near-perfect scores, motivating a decoupled assessment. Conversely, eliminating CoT is also insufficient, since multi-reference evaluation requires explicit logical deduction beyond simple visual matching. As a result, existing MLLM-based evaluators remain caught between text-induced bias and inadequate reasoning capacity.


To tackle these bottlenecks, we present an RL framework for multi-reference image editing with a multi-dimensional Evaluation--erification Reward (EVR). We decompose evaluation into five criteria: reference consistency, scene consistency, visual harmony, instruction consistency, and overall visual quality, to reduce cognitive overload and curb hallucinations. EVR resolves the CoT dilemma by disentangling reasoning from visual grounding: an Evaluator uses a short CoT to produce a rationale and score, preserving necessary deduction while mitigating text-induced bias. Because short-CoT judgments on complex visuals are noisy and can collapse to inconsistent or inflated scores, the Evaluator generates multiple independent hypotheses. While some inevitably include ungrounded hallucinations, this broader pool consistently captures valid visual insights. A Verifier then cross-checks each claim against concrete visual evidence, retaining only evidence-supported assertions and converting them into reliable, visually grounded reward signals. Even when employing the same MLLM, this verification step is highly effective because assessing the validity of a specific factual assertion is a fundamentally simpler and more visually grounded task than conducting an open-ended evaluation. Finally, we build a tailored data pipeline that supplies the required inputs for policy fine-tuning of diffusion models in this multi-reference setting.

Our contributions are summarized as follows:
\begin{itemize}
\item We introduce a reinforcement learning framework for multi-reference image editing, built around a multi-dimensional Evaluation--Verification Reward (EVR). The Evaluator generates multiple hypotheses per criterion, and a Verifier grounds each claim in visual evidence, yielding stable and visually grounded rewards for policy optimization.
\item We design a scalable data pipeline for multi-reference image editing that assembles the full input set needed to fine-tune diffusion models with reinforcement learning under multi-reference conditions.
\item Through extensive experiments on test datasets and real-world examples, we show that our method achieves a 67\% human preference win rate against the base model, substantially improving consistency and visual harmony to match or surpass specialized systems such as NanoBanana.
\end{itemize}

%% file: sections/02_related_work.tex
\section{Related Works}
\subsection{Multi-Reference Image Editing}
The advent of diffusion models\citep{ddpm,flowmatching,Reflow,yu2026elucidating} has established text-to-image (T2I) generation\citep{ldm,MMDiT,flux} as a cornerstone of generative AI, driving rapid advancements in image editing. Early methods relied on inversion-based techniques or explicit controls like Textual Inversion\citep{textualinversion}, ControlNet\citep{controlnet}, and IP-Adapter\citep{ipadapter}. Recent research\citep{anydoor,shiningyourself} has extended these paradigms to multi-reference editing. Furthermore, integrating Diffusion Transformers (DiT)\citep{dit} with Multimodal Large Language Models (MLLMs) has spawned unified models\citep{Qwen-Image,seedream3,seedream4,Nanobanana,flux-kontext} capable of processing multiple references alongside complex instructions for context-aware editing. Despite their generative power, these models still struggle with object consistency\citep{mico}, editing harmony, and aligning with nuanced human judgment standards.
\begin{figure*}[t]
  \centering
  \includegraphics[width=0.98\textwidth]{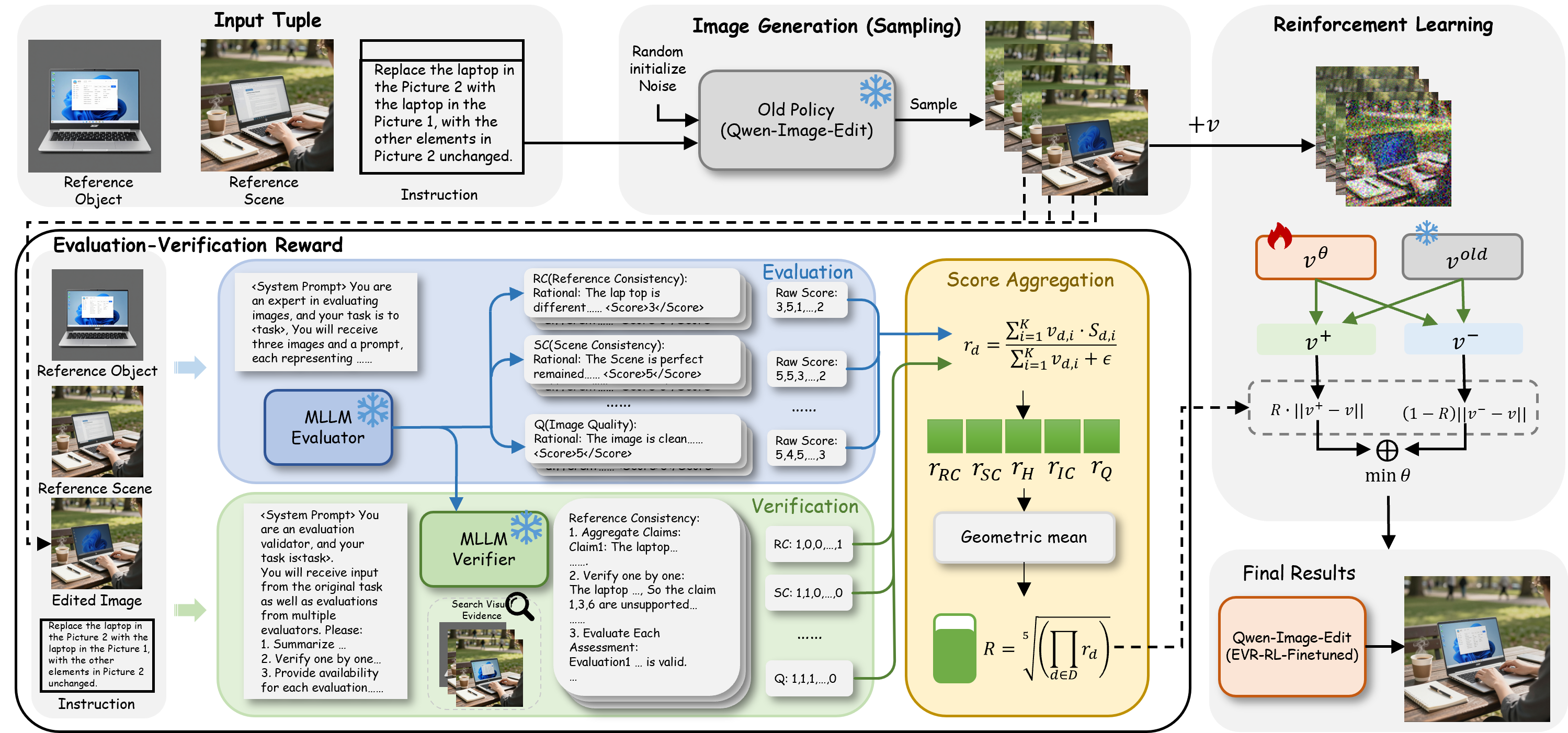}
  \caption{Our process comprises Sampling, EVR, and RL. The EVR module combines MLLM evaluation with visual verification to generate stable rewards, which guide the RL fine-tuning for precise image editing.}
  \label{fig:ppl}
\end{figure*}
\subsection{Reinforcement Learning for Diffusion Models}
The success of Reinforcement Learning (RL) in aligning Large Language Models~\citep{GRPO} has catalyzed similar advancements in Text-to-Image generation and editing~\citep{hpsv2,hpsv3,unifiedreward,imagereward-REFL,clipscore,pickscore}. Early approaches like Diffusion-DPO~\citep{diffusion-dpo} eliminated explicit reward models but remained constrained by static offline datasets, lacking the dynamic feedback essential for iterative editing. Subsequently, Group Relative Policy Optimization (GRPO)~\citep{GRPO,flow-grpo,dancegrpo,mixgrpo,tempflowGRPO} and its variants introduced scalable online learning for diffusion architectures. Recent advancements like DiffusionNFT~\citep{diffusionnft} and AWM~\citep{awm} have further stabilized this online paradigm by directly optimizing the forward process, mitigating training instabilities. However, the efficacy of these advanced RL algorithms is fundamentally bottlenecked by the quality of the reward signal. Applying these RL frameworks to multi-reference editing remains largely underexplored simply because existing reward models fail to provide the reliable, multi-dimensional feedback required to guide the policy toward high-fidelity multi-reference image editing.

\subsection{Reward Modeling with Vision-Language Models}
Vision-language models (VLMs)~\citep{clip} such as CLIP have been widely used as reward sources due to their ability to measure semantic alignment, with extensions like PickScore~\citep{pickscore} refining this via human preference data. However, operating purely in embedding space, they lack the explicit reasoning capabilities required for evaluating complex compositional attributes. While MLLMs (e.g., LLaVA, Qwen-VL~\citep{llava,qwen2-vl,qwen3vl,gpt}) offer a natural-language-based alternative, existing specialized MLLM editing rewards like EditScore~\citep{editscore,unifiedreward,rewarddance} inherit domain biases from narrow, single-image training data. Furthermore, as noted by Uniworld-v2(Edit-r1)\citep{uniworld}, directly employing MLLMs as zero-shot evaluators for multi-image tasks introduces a critical dilemma: relying on lengthy Chain-of-Thought (CoT) reasoning biases the model toward internal textual narratives, causing visual hallucinations; conversely, omitting CoT removes the logical deduction necessary for multi-reference cross-referencing. Consequently, effectively harnessing the reasoning capacity of MLLMs while strictly grounding their evaluations in visual facts remains a critical open challenge in this domain.

%% file: sections/03_method.tex
\section{Method}

To enhance the capabilities of unified image editing models in complex multi-reference scenarios, we propose a reinforcement learning framework driven by a Multi-dimensional Evaluation Verification Reward (EVR) mechanism, as shown in Fig.~\ref{fig:ppl}. As illustrated in Fig.~\ref{fig:dpp}, our approach first utilizes a scalable data pipeline (Sec.~\ref{subsec:data_pipeline}) to construct a dedicated offline dataset of semantically aligned input tuples, each consisting of reference images and text instructions. With these established inputs, we then execute the DiffusionNFT optimization process (Sec.~\ref{subsec:pre}). During each training iteration, the reference policy employs random initialization to explore the generative space and sample a diverse set of candidate edited images for a given input tuple. Subsequently, the EVR mechanism (Sec.~\ref{subsec:evr}) evaluates every generated candidate across five distinct dimensions. By directing an Evaluator to generate multiple rationales and requiring a Verifier to confirm them against concrete visual evidence, we successfully filter out hallucinations and compute a stable and reliable scalar reward. Finally, we optimize the generative policy model to maximize this verified reward using the explicit DiffusionNFT training objective.

\subsection{Preliminary}
\label{subsec:pre}

\paragraph{Flow Matching.} Given a data distribution $p_{\text{data}}({\bm{x}})$, Flow Matching~\citep{flowmatching,Reflow} trains a velocity field ${\bm{v}}_\theta({\bm{x}}_t,t,\bm{c})$ to approximate the conditional probability path from noise ${\bm{x}}_0 \sim \mathcal{N}(0, I)$ to data ${\bm{x}}_1 \sim p_{\text{data}}$, where $\bm{c}$ is the condition and $t$ is the timestep. The generative process is then defined by solving:
\begin{equation}
    \frac{d{\bm{x}}_t}{dt} = {\bm{v}}_\theta({\bm{x}}_t,t, \bm{c}), \quad t \in [0,1],
\end{equation}
Rectified Flow defines a linear flow with $\bm{v}=\bm{x}_0-\bm{x}_1$ as the target velocity field, and the optimization objective becomes: 
\begin{equation}
    \mathcal{L}_{FM} = \mathbb{E}_{t,\bm{x}_0,\bm{x}_1}[||\bm{v}-\bm{v}_\theta(\bm{x}_t,t,\bm{c})||_2^2]
\end{equation}
This enables faster sampling and stable training.

\paragraph{Qwen-Image-Edit.} Our foundational generative framework is Qwen-Image-Edit\citep{Qwen-Image}, an open-source, SOTA model for image generation and editing built upon DiT and FM. By leveraging Qwen2.5-VL\citep{qwen2-vl}, it is capable of processing complex inputs comprising both textual instructions and multiple reference images. 
Specifically, the Qwen2.5-VL first ingests all reference images and text prompts to interpret user intent and generate refined text tokens. Simultaneously, an image encoder is employed to extract visual tokens from each reference image. The text condition $c_t$ and visual condition $c_v$ can be formulated to:
\begin{equation}
    \bm{c}_\text{text}=\mathcal{M}(c_{p},r_1,\dots ,r_n),\  \bm{c}_\text{visual} = [\mathcal{E}(r_1),\dots,\mathcal{E}(r_n)]
\end{equation}
Adopting the MMDiT paradigm, the model concatenates these text and image embeddings prior to the attention mechanism. To maintain spatial coherence across various inputs, Qwen-Image-Edit utilizes MS-RoPE to assign unique positional encodings to each reference image. During the training phase, we exclusively employ LoRA on the transformer parameters to compute policy gradients.

\paragraph{DiffusionNFT} Unlike GRPO-like algorithms, which enable exploration by replacing the ODE with an SDE, DiffusionNFT\citep{diffusionnft} performs exploration through random initialization and defines its policy optimization objective using a reward signal $r(x_0, c)\in[0,1]$ and positive/negative policies constructed from the velocity predictions of the old and new policies:
\begin{equation}
\begin{aligned}
        \mathcal{L}(\theta) &= \mathbb{E}_{\bm{c},\pi^{\text{old}}(\bm{x}_0|\bm{c}),t} \Bigl[ r \left\| \bm{v}^+_\theta(\bm{x}_t, \bm{c}, t) - \bm{v} \right\|_2^2 \\&+ (1 - r) \left\| \bm{v}^-_\theta(\bm{x}_t, \bm{c}, t) - \bm{v} \right\|_2^2 \Bigr]
\end{aligned}
\end{equation}
where $ v $ is the target velocity field. The implicit positive and negative policies $ v^+_\theta $ and $ v^-_\theta $ are combinations of the old policy $ v^{\text{old}} $ and the training policy $ v_\theta $, weighted by a hyperparameter $ \beta $:
\begin{align}
    \bm{v}^+_\theta(\bm{x}_t, \bm{c}, t) &:= (1 - \beta)\, \bm{v}^{\text{old}}(\bm{x}_t, \bm{c}, t) + \beta\, \bm{v}_\theta(\bm{x}_t, \bm{c}, t), \\
    \bm{v}^-_\theta(\bm{x}_t, \bm{c}, t) &:= (1 + \beta)\, \bm{v}^{\text{old}}(\bm{x}_t, \bm{c}, t) - \beta\, \bm{v}_\theta(\bm{x}_t, \bm{c}, t).
\end{align}

\subsection{Scalable Multi-Reference Instruction-Editing Data Pipeline}
\label{subsec:data_pipeline}
\begin{figure}[t]
  \centering
  \includegraphics[width=\linewidth]{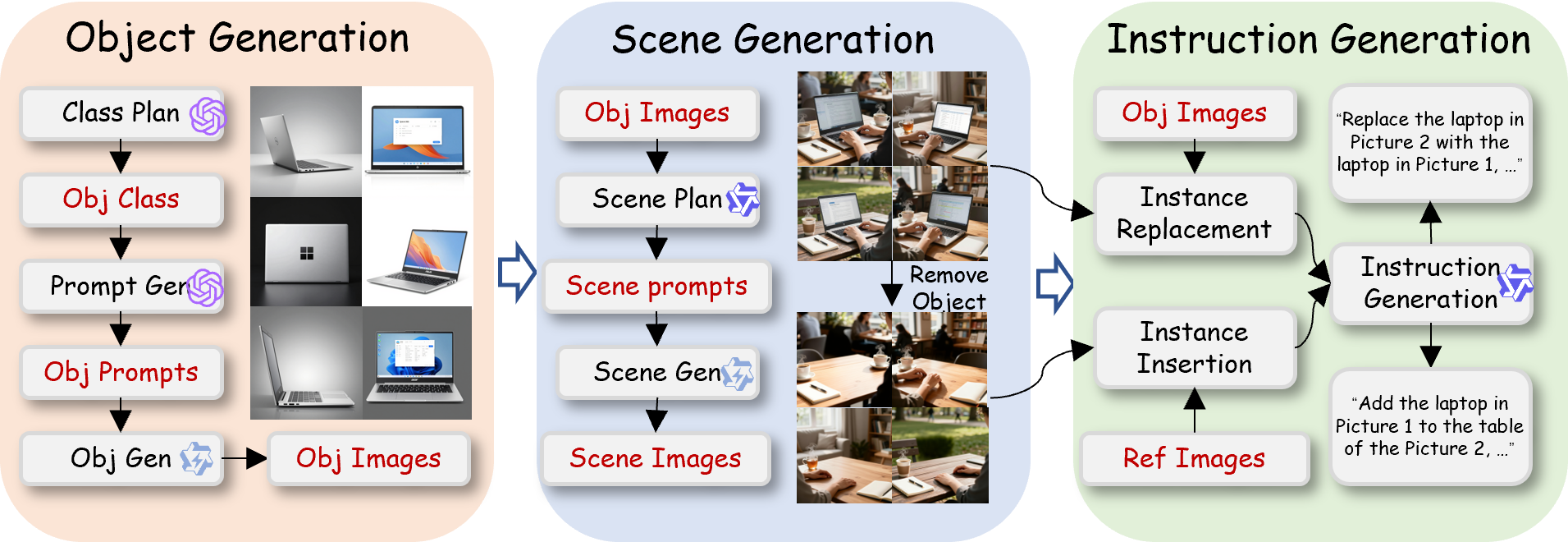}
  \caption{Our pipeline generates Objects, Scenes, and Instructions to create diverse editing pairs covering common daily life categories.}
  \label{fig:dpp}
  
\end{figure}
Existing multi-reference editing datasets rely on real reference and target image pairs~\citep{mico}. Because these pairs are typically collected manually or constructed via video frame segmentation, such approaches are either expensive to scale or inherently prone to visual artifacts. Fortunately, reinforcement learning circumvents this bottleneck by learning purely from the rewards of the model-generated samples. This paradigm only requires semantically coherent input tuples $(\{r_i\}_{i=1}^N, \mathbf{c})$ rather than exact ground truth targets.

However, naively composing these tuples from generic datasets (e.g., Subject200K~\citep{omnicontrol}) often yields severe semantic mismatches. For instance, arbitrarily pairing an image of a car with an indoor bedroom scene creates an illogical instruction that completely derails policy optimization. To resolve this mismatch and unlock highly scalable data generation, we design a task-driven pipeline focused on a representative $N=2$ setting: one foreground reference object $r_{\text{obj}}$ and one background reference scene $r_{\text{scene}}$. By explicitly enforcing strict semantic alignment within these inputs, our pipeline provides reliable and well-posed environments for the RL policy to efficiently optimize object consistency and visual harmony.
The pipeline, as shown in Fig.~\ref{fig:dpp}, operates in three sequential stages:
 
\begin{enumerate}
    \item \textit{Reference Object generation}: We first use GPT\citep{gpt} to generate a diverse set of object categories. For each object category, we further prompt GPT-4 to produce multiple detailed image prompts. These prompts are then fed into Z-Image\citep{z-image} to synthesize reference object images $r_{\text{obj}}$. All generated $r_{\text{obj}}$ are filtered by MLLM-based quality control to ensure visual fidelity and semantic clarity.
    
    \item \textit{Reference Scene generation}: To obtain scene images semantically compatible with $r_{\text{obj}}$, we use MLLM to generate scene prompts conditioned on the object description. Using these prompts, Z-Image synthesizes the original scene image $r_{\text{scene}}$.
    
    \item \textit{Instruction generation}: We define two editing tasks: instance replacement and instance insertion. For instance replacement, we randomly pair reference objects and scenes within the same object category, enabling cross-sample replacement. For instance insertion, we first apply a dedicated inpainting model to remove the target object from $r_{scene}$, producing a clean background $r'_{scene}$; we then pair this background with a reference object $r_{\text{obj}}$ from the same category to form an insertion task. In both cases, Qwen3-VL\citep{qwen3vl} generates grounded natural-language instructions $p$ that precisely describe the desired edit.
\end{enumerate}

The final dataset comprises 10K such triplets $(r_{\text{obj}}, r_{\text{scene}}, p)$. From this collection, we sample 1K triplets to form a test set, which simultaneously validates our reward model and assesses the overall editing quality. Additionally, we collect over 300 real-world multi-reference editing examples from the internet. This external set is utilized to evaluate out-of-domain generalization and the ability of the model to handle an arbitrary number of reference images beyond the $N=2$ training setting.

\subsection{Multi-Dimensional Evaluation-Verification Reward}
\label{subsec:evr}
A core challenge in reinforcement learning for image editing is designing a reliable reward function. To comprehensively evaluate the results of multi-reference image editing and to mitigate the inherent hallucination noise of language model reasoning, we propose the Multi-Dimensional Evaluation Verification Reward (EVR) mechanism. Specifically, EVR explicitly decomposes the reward process into three distinct stages: rationale-based evaluation across multiple dimensions, visual grounding verification, and geometric aggregation.

\subsubsection{Multi-Dimensional Evaluation with Rationales}
Given an edited image $\hat{I}$, its reference images (object reference $r_{\text{obj}}$ and scene reference $r_{\text{scene}}$), and the editing instruction $p$, we task an MLLM Evaluator $E$ (e.g., Qwen3-VL-8B-Instruct) to perform a structured assessment. Evaluating complex edits like the bottle insertion in Fig.~\ref{fig:qualitative3} requires assessing both identity consistency and physical plausibility, such as finger occlusion, which single scalar scores cannot capture. Therefore, $E$ evaluates the image across five key dimensions:
\begin{itemize}
    \item \textbf{Reference Consistency (RC):} Fidelity of the edited object to $r_{\text{obj}}$.
    \item \textbf{Scene Consistency (SC):} Fidelity of the target scene to $r_{\text{scene}}$.
    \item \textbf{Harmony (H):} Visual harmony between the object and the scene.
    \item \textbf{Instruction Consistency (IC):} Success in executing the instruction $p$.
    \item \textbf{Visual Quality (Q):} Absence of artifacts and overall aesthetic appeal.
\end{itemize}

For each dimension $d \in D$, the evaluator is designed to perform $K$ independent evaluations. Each evaluation $i \in \{1, \dots, K\}$ consists of a natural-language rationale $T_{d,i}$ and a numerical score $S_{d,i} \in [1, 5]$:
\begin{equation}
    \{(T_{d,i}, S_{d,i})\}_{i=1}^K = E(r_{\text{obj}}, r_{\text{scene}}, p, \hat{I}, d)
\end{equation}

\subsubsection{Visual Grounding Verification}
To suppress hallucinated feedback, we introduce an MLLM-based Verifier $V$. For each editing capability dimension $d$, $V$ first consolidates all evaluator-provided rationales $\{T_{d,i}\}_i$ into a structured set of verifiable claims. It then inspects the generated image $\hat{I}$, along with the reference images $r_{\text{obj}}$, scene $r_{\text{scene}}$, and editing instruction $p$, to determine whether each claim is visually grounded in the actual pixels. Based on this verification, $V$ assigns a reliability sign $v_{d,i} \in \{0,1\}$ to each specific rationale $i$, reflecting whether this rationale is supported by evidence in $\hat{I}$:
\begin{equation}
    \{v_{d,i}\}_{i=1}^K = V(r_{\text{obj}}, r_{\text{scene}}, p, \hat{I}, \{T_{d,i}\}_{i=1}^K)
\end{equation}

\subsubsection{Reward Aggregation}
The verified score for each dimension $r_d$ is calculated as the weighted average of the scores, ensuring that only grounded evaluations contribute to the learning signal. The final reward $R$ is then computed using the geometric mean of all dimension rewards:
\begin{equation}
    r_d = \frac{\sum_{i=1}^K v_{d,i} \cdot S_{d,i}}{\sum_{i=1}^K v_{d,i} + \epsilon}, \quad R = \sqrt[5]{\prod_{d \in \{RC, SC, H, IC, Q\}} r_d}
\end{equation}
where $\epsilon$ is a small constant for numerical stability. Following the common practice\citep{editscore}, we use the geometric mean rather than the arithmetic mean to prevent reward hacking, where a model might significantly improve one dimension (e.g., Visual Quality) while completely failing another (e.g., Instruction Consistency). This multiplicative structure ensures that a high total reward is only achievable if the model performs well across all facets of the multi-reference editing task.
\begin{figure}[t]
  \centering
  \includegraphics[width=\linewidth]{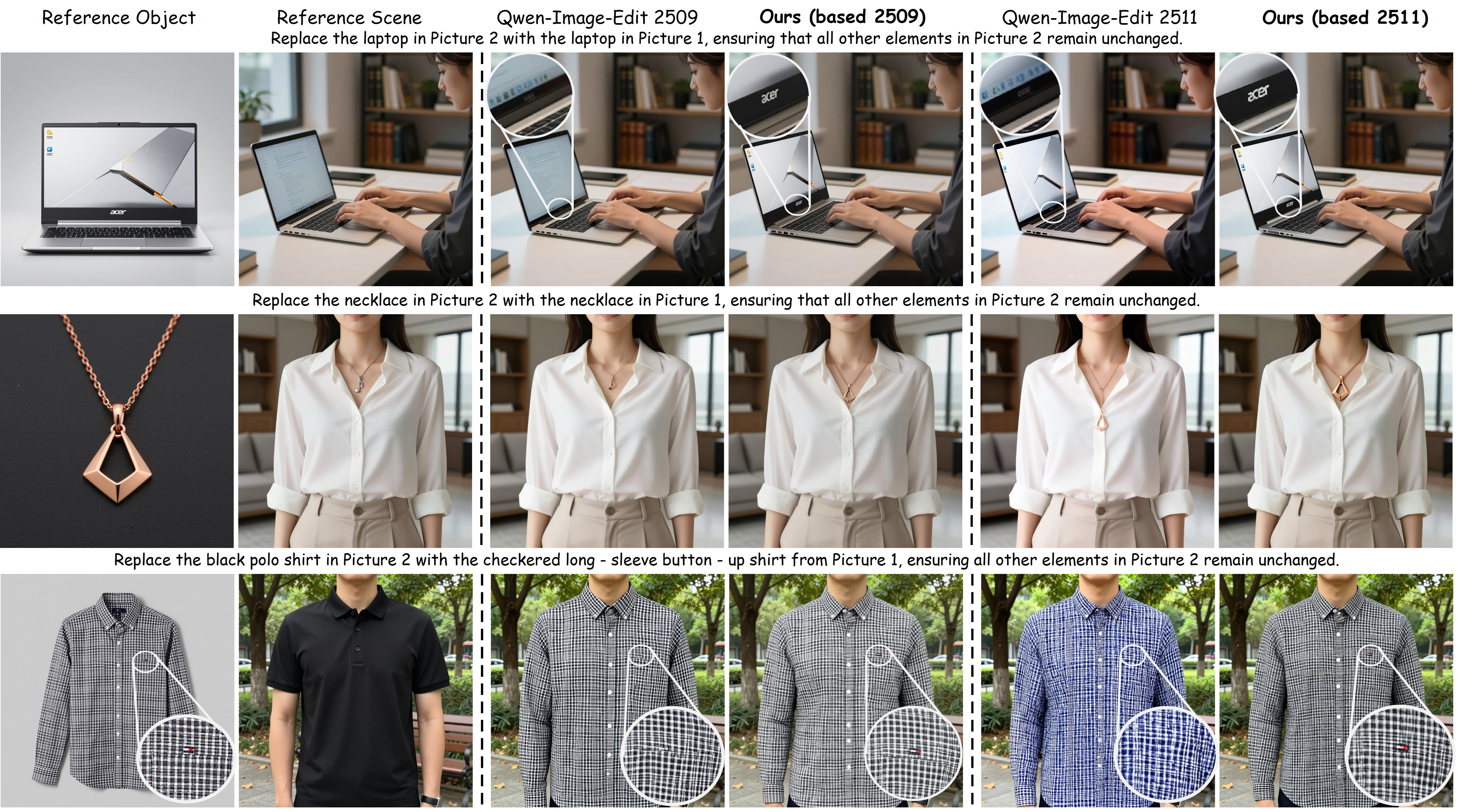}
  \caption{Qualitative comparison between baseline models and our method. Our method demonstrates superior reference consistency, particularly in preserving fine-grained structural details and subtle textures that baseline models frequently distort or omit. }
  \label{fig:qualitative1}
\end{figure}
\begin{figure}[t]
    \centering
    \includegraphics[width=\linewidth]{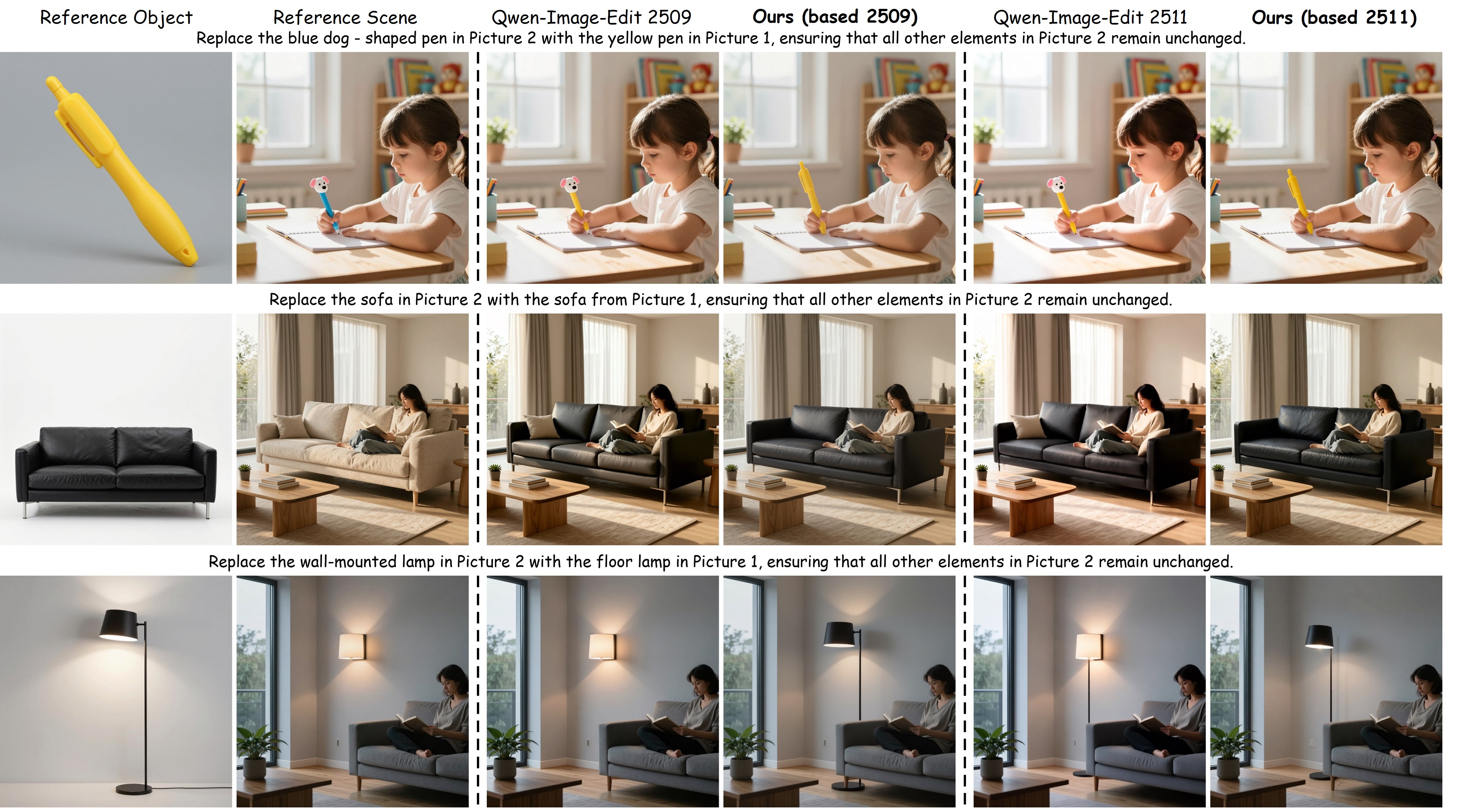}
    \caption{Qualitative comparison between baseline models and our method. Our method demonstrates a superior ability for context disambiguation. The base models cannot distinguish whether the dog decoration, three-seat cushions, and transparent lampshade are features of objects or scenes.}
    \label{fig:qualitative2}
\end{figure}

%% file: sections/05_experiments.tex
\section{Experiments}

\subsection{Experimental Setup}
\begin{figure}[t]
    \centering
    \includegraphics[width=\linewidth]{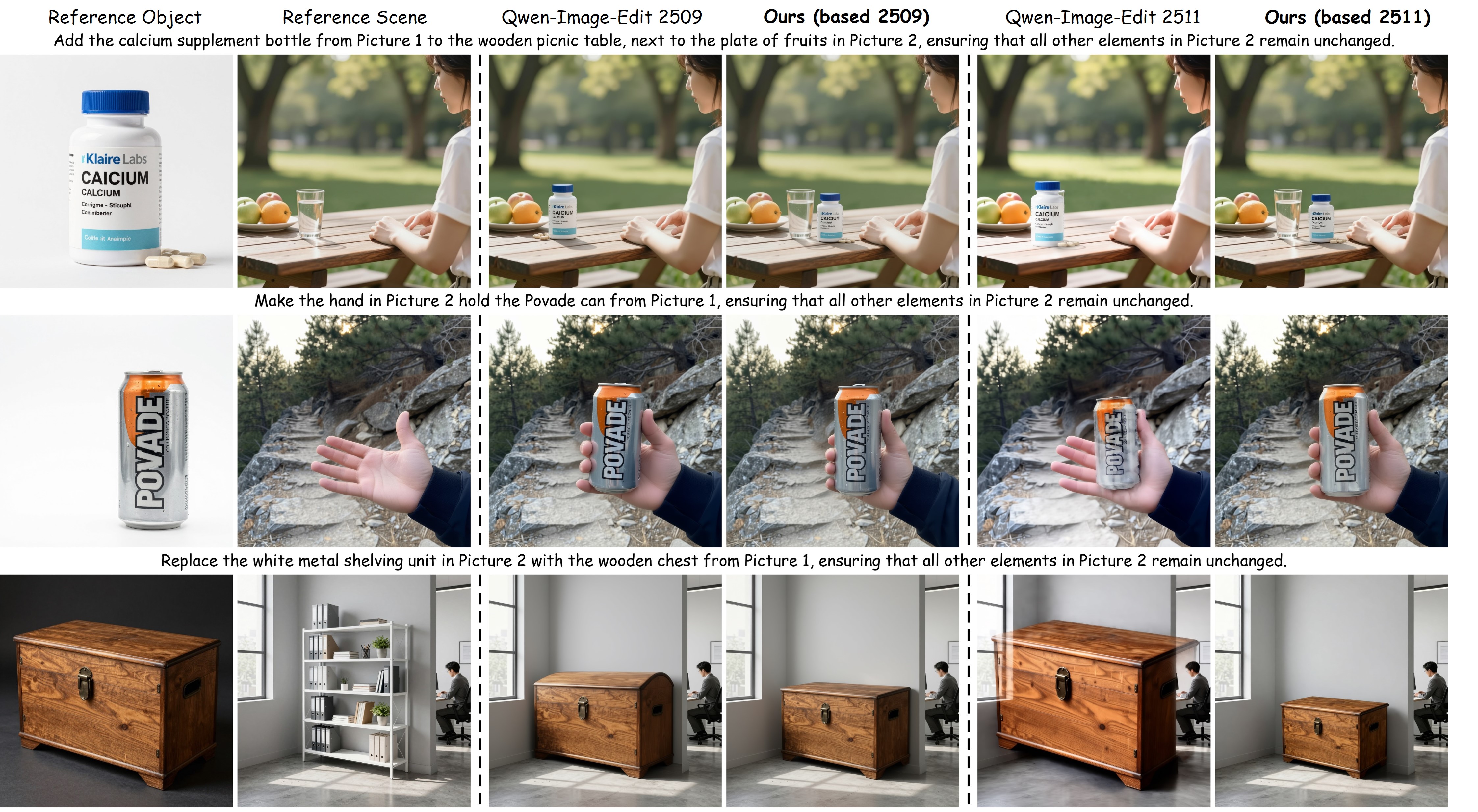}
    \caption{Qualitative comparison between Qwen-Image-Edit versions and our RL-tuned model. Our method shows superior harmony and instruction consistency.}
    \label{fig:qualitative3}
\end{figure}
\begin{table*}[t]
    \centering
    \fontsize{7pt}{8pt}\selectfont
    \caption{Quantitative comparison of multi-reference editing performance. For greater persuasiveness and to better explain that the benefits are not due to reward hacking, the metrics are calculated using the qwen3.5-plus API.}
    \label{tab:main_results}
    \resizebox{\textwidth}{!}{%
    \begin{tabular}{lllllll}
    \toprule
    Model   & RC $\uparrow$         & SC     $\uparrow$    & H $\uparrow$     & IC  $\uparrow$  & Q $\uparrow$    & Score  $\uparrow$     \\ \hline
    Qwen-Image-Edit-2509  & 3.35 & {4.62}& 2.22  & 3.27 & 4.18& 0.56 \\ 
    +logit Reward(Edit-R1/Uniworld-v2)  & 4.02\textcolor{ForestGreen}{(+0.77)}  &  4.55\textcolor{Gray}{(-0.07)}    & 2.54\textcolor{ForestGreen}{(+0.32)}   &  3.83\textcolor{ForestGreen}{(+0.56)}   & 4.23\textcolor{Gray}{(+0.05)}    &0.65\textcolor{ForestGreen}{(+0.09)} \\
    \textbf{+EVR-RL(Qwen3-VL-8B-Instruct)}  & {{4.36\textcolor{ForestGreen}{(+1.01)}}}    & {4.51\textcolor{Gray}{(-0.11)}}          & {{2.63\textcolor{ForestGreen}{(+0.41)}}}    & {3.98\textcolor{ForestGreen}{(+0.72)}}          & {4.15\textcolor{Gray}{(-0.03)}}          & {{0.68\textcolor{ForestGreen}{(+0.12)}}}    \\
    \textbf{+EVR-RL(Qwen3-VL-32B-Instruct)} & {4.27\textcolor{ForestGreen}{(+0.93)}}          & {4.41\textcolor{Gray}{(-0.21)}}          & {2.57\textcolor{ForestGreen}{(+0.35)}}          & {\textbf{4.03\textcolor{ForestGreen}{(+0.77)}}} & {{4.18\textcolor{Gray}{(+0.00)}}}    & {0.68\textcolor{ForestGreen}{(+0.12)}}          \\
    \textbf{+EVR-RL(Qwen3.5-plus)}          & {\textbf{4.43\textcolor{ForestGreen}{(+1.08)}}} & {\textbf{4.67\textcolor{Gray}{(+0.05)}}} & {\textbf{2.73\textcolor{ForestGreen}{(+0.51)}}} & {{4.01\textcolor{ForestGreen}{(+0.75)}}}    & {\textbf{4.18\textcolor{Gray}{(+0.00)}}} & {\textbf{0.70\textcolor{ForestGreen}{(+0.14)}}} \\\hline
    Qwen-Image-Edit-2511                    & 3.66                                        & 4.43                                        & 2.25                                        & 3.57                                        & 4.18                                        & 0.59                                        \\
    \textbf{+EVR-RL(Qwen3-VL-8B-Instruct)}  & {\textbf{4.37\textcolor{ForestGreen}{(+0.71)}}} & {\textbf{4.63\textcolor{ForestGreen}{(+0.20)}}} & {\textbf{2.60\textcolor{ForestGreen}{(+0.35)}}} & {\textbf{4.12\textcolor{ForestGreen}{(+0.55)}}} & {\textbf{4.21\textcolor{Gray}{(+0.03)}}} & {\textbf{0.70\textcolor{ForestGreen}{(+0.11)}}}\\\bottomrule
    \end{tabular}%
    }
\end{table*}
\paragraph{Implementation Details.} 
We adopt Qwen-Image-Edit as our base model. For the policy optimization, we employ Low-Rank Adaptation (LoRA)~\citep{lora} with a rank of 64 on the DiT blocks. We use the AdamW\citep{AdamW} optimizer with a learning rate of $5 \times 10^{-5}$. \textbf{With the exception of Fig.~\ref{fig:cross} and the model robustness experiments, we employ Qwen3-VL-8B-Instruct as both the Evaluator and the Verifier for the EVR mechanism.} Each training epoch processes 24 distinct input tuples, with 8 stochastic samples generated per condition (i.e., 192 edited images per epoch). Training is conducted for 40 epochs on 4 NVIDIA H20 GPUs, totaling approximately 150 GPU-hours. We use vLLM~\citep{kwon2023efficient} to deploy the MLLM. Other settings follow Uniworld-v2(Edit-r1).\citep{uniworld}. We evaluate models across five dimensions: Reference Consistency (RC), Scene Consistency (SC), Harmony (H), Instruction Consistency (IC), and Visual Quality (Q). Due to cost constraints, except for the metrics in main results which are calculated using the qwen3.5-plus API, all other metrics are calculated using the locally deployed Qwen3-VL-8B-Instruct. In addition to our proposed EVR scores, we report results from human expert evaluations. 

\subsection{Main Results}
\subsubsection{Comparison with base model}
We compare our RL-tuned model against base Qwen-Image-Edit (v2509, v2511). As shown in Table.~\ref{tab:main_results}, our method yields substantial gains in Reference Consistency, Harmony, and Instruction Consistency. The slight dip in Scene Consistency is expected, as baseline models spuriously inflate scores by simply copying the reference scene image. This collapse mode, visually evident in Fig.~\ref{fig:qualitative1} (top row) and Fig.~\ref{fig:qualitative2} (bottom row), also matches the findings from our w/o Dim ablation. Our approach prevents this shortcut to ensure genuine editing. Meanwhile, Image Quality remains stable, as the base model is already strong; we employ quality rewards primarily as a regularizer against reward hacking rather than for further optimization.

We also show the qualitative comparison in Fig.~\ref{fig:qualitative1}, \ref{fig:qualitative2}, and \ref{fig:qualitative3}, we highlight four critical improvements achieved by our model:

\textbf{Reference Consistency.} Unlike the base model which often loses fine-grained textures (e.g., logos, screens), our method precisely retains intricate structural details of reference objects (Fig.~\ref{fig:qualitative1}).

\textbf{Context Disambiguation.} We effectively eliminate context leakage where background attributes bleed into objects. Our model strictly separates reference and scene features, enabling accurate replacement of similar-looking items (e.g., handbags, lamps) without semantic confusion (Fig.~\ref{fig:qualitative2}).

\textbf{Instruction Consistency.} Our model demonstrates superior precision in complex tasks, correctly distinguishing between \textit{addition} and \textit{replacement} instructions. It avoids common baseline errors like misinterpreting commands or generating floating artifacts (Fig.~\ref{fig:qualitative3}, top).

\textbf{Editing Harmony.} Even with significant pose or perspective disparities, our approach ensures seamless integration. Newly added objects naturally align with scene lighting and geometry, avoiding the disharmony and misalignment frequent in baseline outputs (Fig.~\ref{fig:qualitative3}, bottom).

\subsubsection{Comparison with Other MLLM-based Reward}
We also evaluate our approach against other MLLM-based reward models. Notably, while Edit-R1 was originally validated exclusively on single-image editing, we adapt its logit-weighting mechanism for comparison within our framework. As analyzed in Section~\ref{sec:reward_validation} and the Appendix, Edit-R1's reward distribution exhibits characteristics similar to direct scoring, yet demonstrates enhanced numerical stability. However, this approach yields uninterpretable reward signals and performs poorly on evaluation dimensions that require explicit logical deduction, as it lacks a transparent rationale-to-verification pipeline to suppress text-induced hallucinations. As shown in Table.~\ref{tab:main_results} and Fig.~\ref{fig:cross}, logit-based reward don't excel in tasks requiring reasoning.

\subsubsection{Robustness to Model}
As shown in Table~\ref{tab:main_results} and Fig.~\ref{fig:cross}, the editing quality of our framework scales positively with the intrinsic capability of the MLLM evaluator: stronger models naturally yield higher fidelity due to enhanced visual grounding. Crucially, the EVR mechanism itself exhibits strong robustness across diverse scales. It consistently delivers substantial gains over unverified baselines, particularly on dimensions requiring logical reasoning or prone to hallucination. Most importantly, this verification pipeline unlocks a key advantage: small-scale evaluators equipped with EVR consistently outperform stronger models relying on direct scoring. By explicitly filtering ungrounded claims, EVR effectively compensates for the limited capacity of smaller models, enabling them to generate more reliable reward signals than larger, unverified counterparts.

\subsection{Reward Model Validation}
\label{sec:reward_validation}
\begin{figure}[t]
    \centering
    \includegraphics[width=\linewidth]{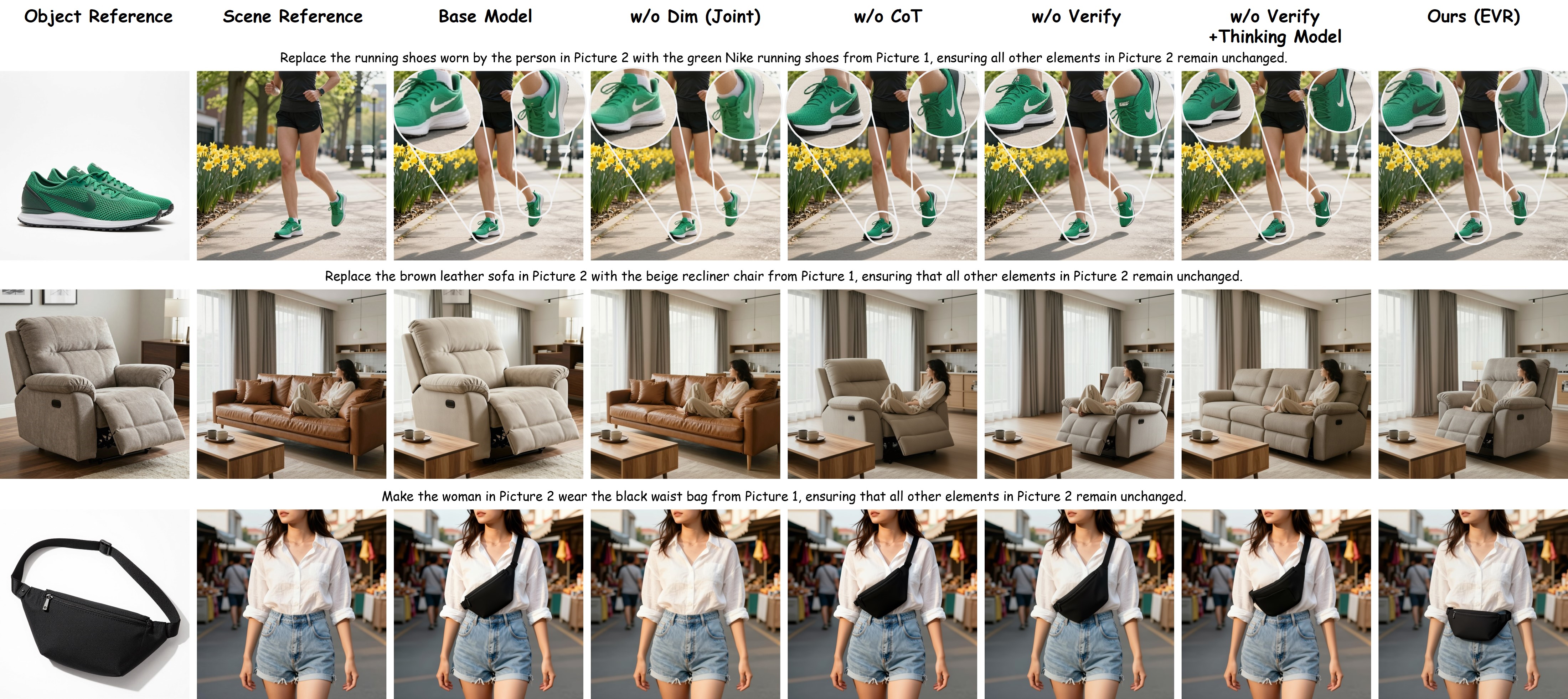}
    \caption{The qualitative demonstration of the ablation study confirms the necessity of our proposed method.}
    \label{fig:ablation}    
\end{figure}

We validate our EVR paradigm against five baseline reward formulations: Joint CoT (evaluating all dimensions comprehensively in a single pass to directly output the final results), Decoupled Direct (scoring each dimension directly without rationales), Decoupled CoT (scoring each dimension via reasoning), CoT Averaging (averaging the scores from five independent reasoning trials), and Logit Weighted (applying the Uniworld-v2(Edit-r1)~\citep{uniworld} strategy to weight the logits of discrete scores). 
As Table~\ref{Tab.diffrewd} shows, our method achieves superior alignment with human judgment. By combining decoupled assessment with strict visual verification, our approach mitigates severe textual hallucinations inherent to pure CoT methods while preserving logical deduction. See Appendix B.1 for reward distributions and qualitative analyses.
\begin{table*}[t]
    \centering
    \scriptsize
    \setlength{\tabcolsep}{3pt}
    \renewcommand{\arraystretch}{1.05}
    \begin{minipage}[t]{0.42\textwidth}
    \centering
    \caption{Human Alignment of different rewards.}
    \label{Tab.diffrewd}
    \begin{tabular*}{\linewidth}{@{\extracolsep{\fill}}lc@{}}
    \toprule
    Reward & Human Align $\uparrow$ \\
    \midrule
    Joint CoT & 0.473 \\
    Decoupled Direct & 0.578 \\
    Decoupled CoT & 0.629 \\
    CoT Averaging & 0.659 \\
    Logit Weighted & 0.643 \\
    \midrule
    EVR (Ours) & \textbf{0.707} \\
    \bottomrule
    \end{tabular*}
    \end{minipage}
    \hfill
    \begin{minipage}[t]{0.54\textwidth}
    \centering
    \caption{Quantitative analysis of ablation.}
    \label{Tab.Ablation}
    \setlength{\tabcolsep}{2.4pt}
    \begin{tabular*}{\linewidth}{@{\extracolsep{\fill}}lcccccc@{}}
    \toprule
    Config & RC & SC & H & IC & Q & Score \\
    \midrule
    Base & 3.35 & 4.37 & 3.34 & 3.66 & 4.46 & 0.66 \\
    w/o Dim & 1.74 & \textbf{4.86} & 1.41 & 1.42 & \textbf{4.48} & 0.30 \\
    w/o CoT & 3.93 & 4.34 & 3.85 & 4.15 & 4.46 & 0.75 \\
    w/o Veri. & 4.10 & 4.27 & 3.99 & 4.17 & 4.46 & 0.77 \\
    \quad +Thinking & 3.86 & 4.29 & 3.81 & 4.05 & 4.46 & 0.74 \\
    EVR Reward & \textbf{4.15} & 4.25 & \textbf{4.32} & \textbf{4.23} & 4.41 & \textbf{0.79} \\
    \bottomrule
    \end{tabular*}
    \end{minipage}
\end{table*}

\begin{figure}[t]
  \centering
  \includegraphics[width=\linewidth]{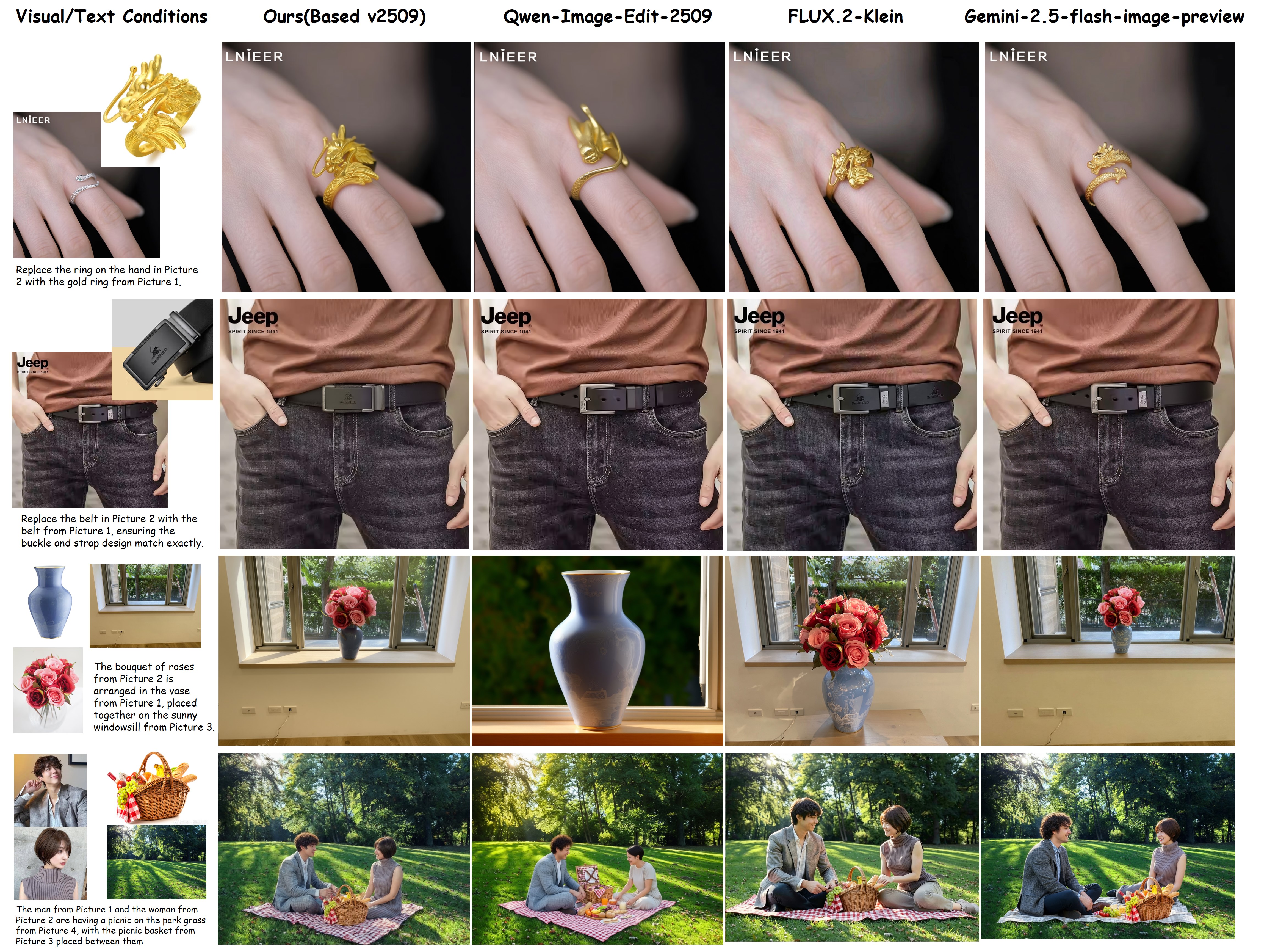}
  \caption{\textbf{Generalization on OOD and N > 2 tasks.} Qualitative comparisons demonstrate that our method outperforms baselines in preserving fine-grained details, unseen categories and complex scenarios.}
  \label{fig:ood}
  \vspace{-1.5em}
\end{figure}
\subsection{Ablation and Generalization}

\paragraph{Ablation Study.} We conducted ablation study based on v2509 to explore the role of different components in the reward model. As shown in Table.~\ref{Tab.Ablation} and Fig.~\ref{fig:ablation}, None of the ablation settings can preserve fine details, particularly when the reference object and the scene exhibit high visual similarity, as demonstrated in the top row of Fig.~\ref{fig:ablation}. Removing the Verifier leads to a notable drop in metrics requiring complex reasoning and lacking direct visual grounding, as the system loses its ability to validate logical coherence, despite maintaining performance on directly observable attributes. For example, in the bottom row of Fig.~\ref{fig:ablation}, the model can not recognize that this is a waist bag.

Replacing the evaluator with a Thinking model degrades performance. Consistent with findings in Uniworld-v2(Edit-r1), excessively long CoT tends to mislead the model with textual priors, causing it to neglect actual image content, as shown in the middle row of Fig.~\ref{fig:ablation}. Conversely, removing CoT entirely harms tasks demanding logical deduction, confirming the necessity of balanced reasoning.

Most critically, replacing our dimension-wise evaluation with a joint judgment results in complete failure: the policy collapses into simply copying the background scene. This manifests as artificially inflated Scene Consistency scores but a total loss of editing capability, highlighting that fine-grained rewards are essential to prevent such reward hacking.
\paragraph{Generalization to the Wild and $N>2$.}
Although our model was trained exclusively on synthetic data with $N=2$, we evaluate its generalization capability on a real-world Out-of-Distribution (OOD) dataset collected from the internet, which includes diverse input tuples with $N\geq2$ input images. As shown in Fig.~\ref{fig:ood}, \ref{fig:user}, we compare our method against the base model, Flux2, and Gemini2.5-Flash-Image (NanoBanana). Both qualitative comparisons and user study results demonstrate that our approach achieves superior performance across all settings on this OOD data. 
Notably, it maintains robust instruction following and visual harmony even for $N>2$ scenarios unseen during training. These results confirm that our method effectively generalizes beyond its training distribution to handle complex, multi-reference editing in the wild.

\subsection{User Study}
\begin{wrapfigure}{r}{0.34\linewidth}
    \centering
    \vspace{-1.0em}
    \includegraphics[width=0.96\linewidth]{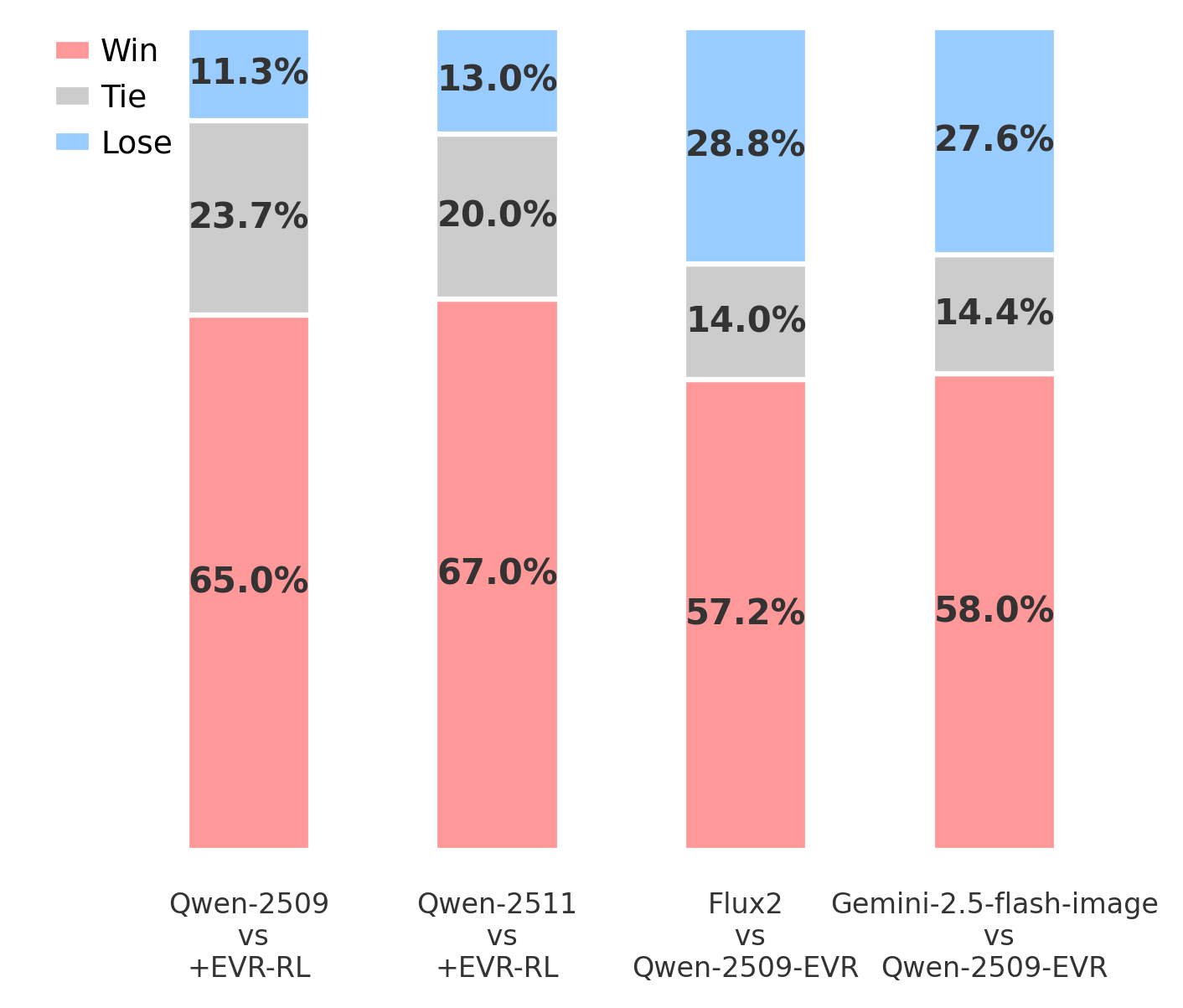}
    \vspace{-0.8em}
    \caption{Results of User Study.}
    \label{fig:user}
    \vspace{-0.3em}
\end{wrapfigure}
We conducted a large-scale user study collecting 6,768 feedback responses to compare our fine-tuned models against their baselines, as well as against leading open-source and closed-source methods. Fig.~\ref{fig:user} demonstrates that our approach significantly improves upon the baseline and achieves superior performance in consistency and harmony for multi-reference image editing compared to competing methods. Detailed statistical results are provided in Appendix B.2.


%% file: sections/07_conclusion.tex
\section{Conclusion}

We propose a reinforcement learning framework for multi-reference image editing that leverages an Evaluation-Verification reward mechanism to suppress hallucinations in MLLM-based feedback. By grounding evaluator rationales in visual evidence, our approach enables stable and human-aligned policy updates without requiring ground-truth edits. Combined with a scalable synthetic data pipeline, our method significantly enhances reference consistency, context disambiguation, instruction consistency, and visual harmony in Qwen-Image-Edit. Experiments show strong performance on both in-domain benchmarks and real-world out-of-domain scenarios, demonstrating its potential for professional-grade image editing.

%% file: sections/appendix/a_implementation_details.tex
\section{Implementation Details }

\subsection{Setting}
We present key hyperparameters of the training in Table.~\ref{tab:set}.
The categories of objects included in the dataset are shown in the Fig.~\ref{fig:ca}

\begin{table*}[ht]
    \centering
    \caption{Implementation Details}
    \label{tab:set}
    \begin{tabular}{c c}
        \toprule
        \textbf{Parameter} & \textbf{Setting} \\
        \midrule
        \multicolumn{2}{c}{\textit{Basic}} \\
        \midrule
        Learning Rate & 5e-5 \\
        $\beta_1$ & 0.9 \\
        $\beta_2$ & 0.999 \\
        Batch Size & 4 \\
        EMA Decay & 0.9 \\
        \midrule
        \multicolumn{2}{c}{\textit{Sampling}} \\
        \midrule
        Sampling Inference Steps & 6 \\
        Resolution & $512 \times 512$ \\
        The Number of Images Per Prompt & 8 \\
        The Number of Groups & 24 \\
        \midrule
        \multicolumn{2}{c}{\textit{DiffusionNFT}} \\
        \midrule
        KL Loss Weight & 0.0001 \\
        Guidance Strength ($\frac{1}{\beta}$) & 1.0 \\
        \midrule
        \multicolumn{2}{c}{\textit{Evaluation-Verification Reward}} \\
        \midrule
        $K$ & 5 \\
        Normalization $R\in[1,5]$ to $[0,1]$ & $R_{final} = \frac{R-1}{4}$\\
        \midrule
        \multicolumn{2}{c}{\textit{Test}} \\
        \midrule
        Sampling Inference Steps & 20 \\
        Resolution & $1024\times1024$ or the resolution of OOD images\\
        \bottomrule
    \end{tabular}
\end{table*}

We set the number of candidate rationales and scores output by the evaluator to five to strike a balance between the verifier's availability and computational overhead. Upon verifier failure, we adopt a fallback strategy of using the mean candidate score to maintain reward stability.

\subsection{\texorpdfstring{Ablation on the Number of Evaluation Hypotheses ($K$)}{Ablation on the Number of Evaluation Hypotheses (K)}}
\label{sec:ablation_k}

To determine the number of evaluation hypotheses ($K$), we conducted an ablation study on 500 generated samples, assessing the top-$n$ availability rate, defined as the probability of obtaining at least one visually grounded and valid evaluation claim. As shown in Table~\ref{tab:k_ablation}, the empirical availability increases from 70\% at $n=1$ to 89\% at $n=5$. Notably, the observed growth rate indicates that these multiple evaluations do not satisfy the assumption of independent and identically distributed (i.i.d.) trials. Instead, the MLLM Evaluator exhibits a systematic bias rather than purely random noise: when struggling with a complex edit, it tends to repeat similar hallucinated textual priors across independent generations. This non-i.i.d.\ phenomenon strongly validates our EVR design: simply averaging evaluations is inherently flawed due to this mode collapse, making our visual Verifier essential for extracting valid reward signals.

\begin{table*}[ht]
\centering
\begin{minipage}[t]{0.42\textwidth}
\centering
\caption{The empirical top-$n$ availability rate of evaluation hypotheses.}
\label{tab:k_ablation}
\begin{tabular}{lc}
\toprule
\textbf{Top-$n$ Hypotheses} & \textbf{Availability Rate} \\
\midrule
$n=1$ & 70\% \\
$n=2$ & 78\% \\
$n=3$ & 82\% \\
$n=4$ & 86\% \\
$n=5$ & 89\% \\
\bottomrule
\end{tabular}
\end{minipage}
\hfill
\begin{minipage}[t]{0.48\textwidth}
\centering
\caption{Reward Statistics Across Different Models}
\label{tab:reward_stats}
\begin{tabular}{lcc}
\toprule
Model & Mean & Variance \\
\midrule
EVR & 0.63 & 0.073  \\
Multi-Direct & 0.79 & 0.043  \\
Multi-CoT & 0.79 & 0.074 \\
Multi-Logit & 0.80 & 0.038 \\
Average & 0.80 & 0.064 \\
Direct Scoring & 0.95 & 0.029  \\
\bottomrule
\end{tabular}
\end{minipage}
\end{table*}

Despite the upward trend in availability, we restrict the maximum $K$ to 5. Increasing $K$ further would linearly scale the computational overhead during the online reinforcement learning phase. More importantly, a larger $K$ necessitates feeding more claims into the Verifier simultaneously, which significantly expands its Chain-of-Thought (CoT) length. As discussed in the main text, overly long CoT contexts overwhelm the MLLM, exacerbating text-induced bias and degrading its visual grounding capabilities. Therefore, $K=5$ serves as a trade-off between reward stability and verification reliability.

%% file: sections/appendix/b_configuration.tex
\section{Additional Analysis and Results}

\subsection{Reward Distributions and qualitative analyses}
\begin{figure*}[ht]
    \centering
    \begin{minipage}[t]{0.52\textwidth}
        \centering
        \includegraphics[width=\linewidth]{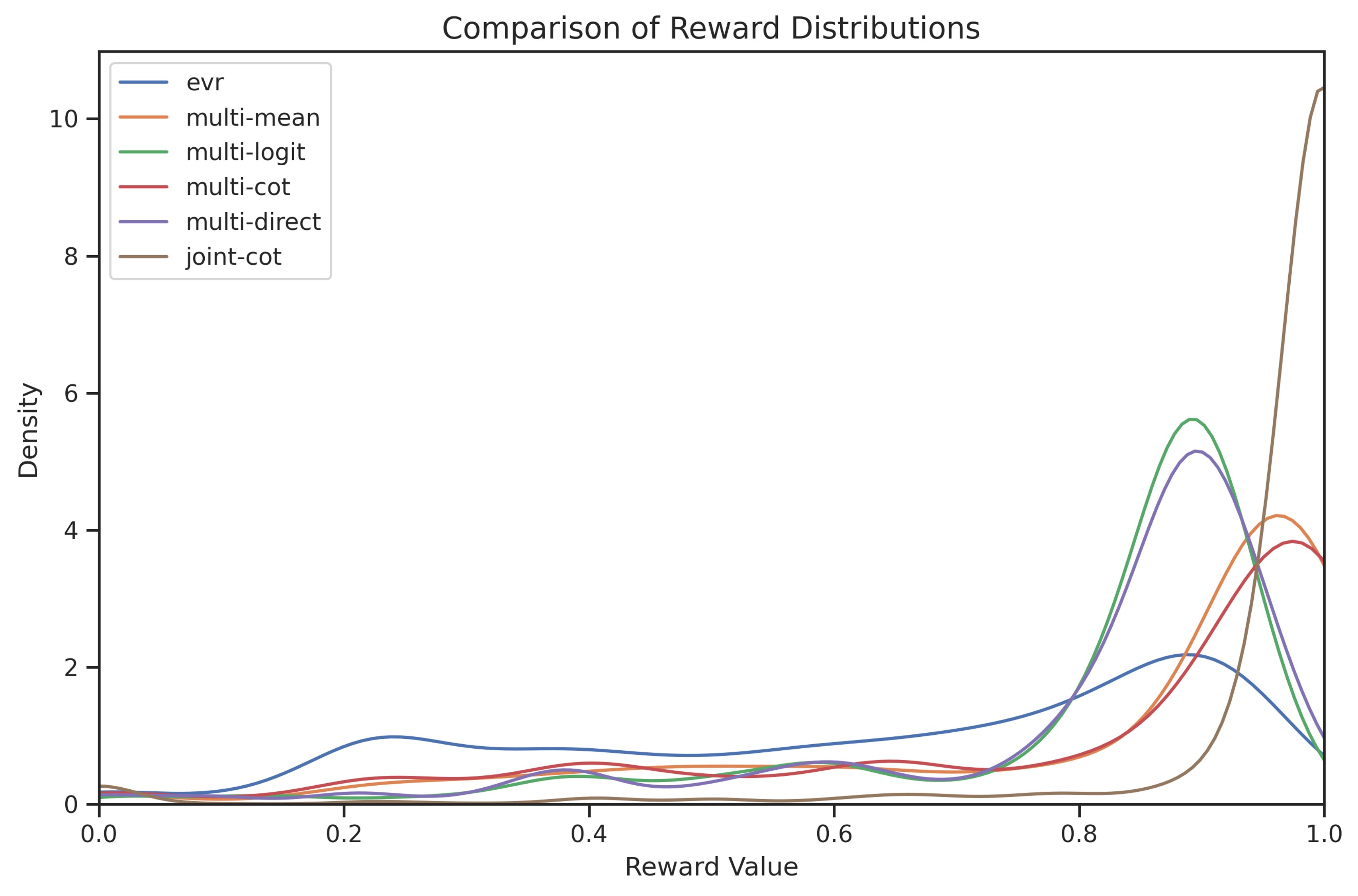}
        \caption{Distribution Curve.}
        \label{fig:reward_dist}
    \end{minipage}
    \hfill
    \begin{minipage}[t]{0.42\textwidth}
        \centering
        \includegraphics[width=0.92\linewidth]{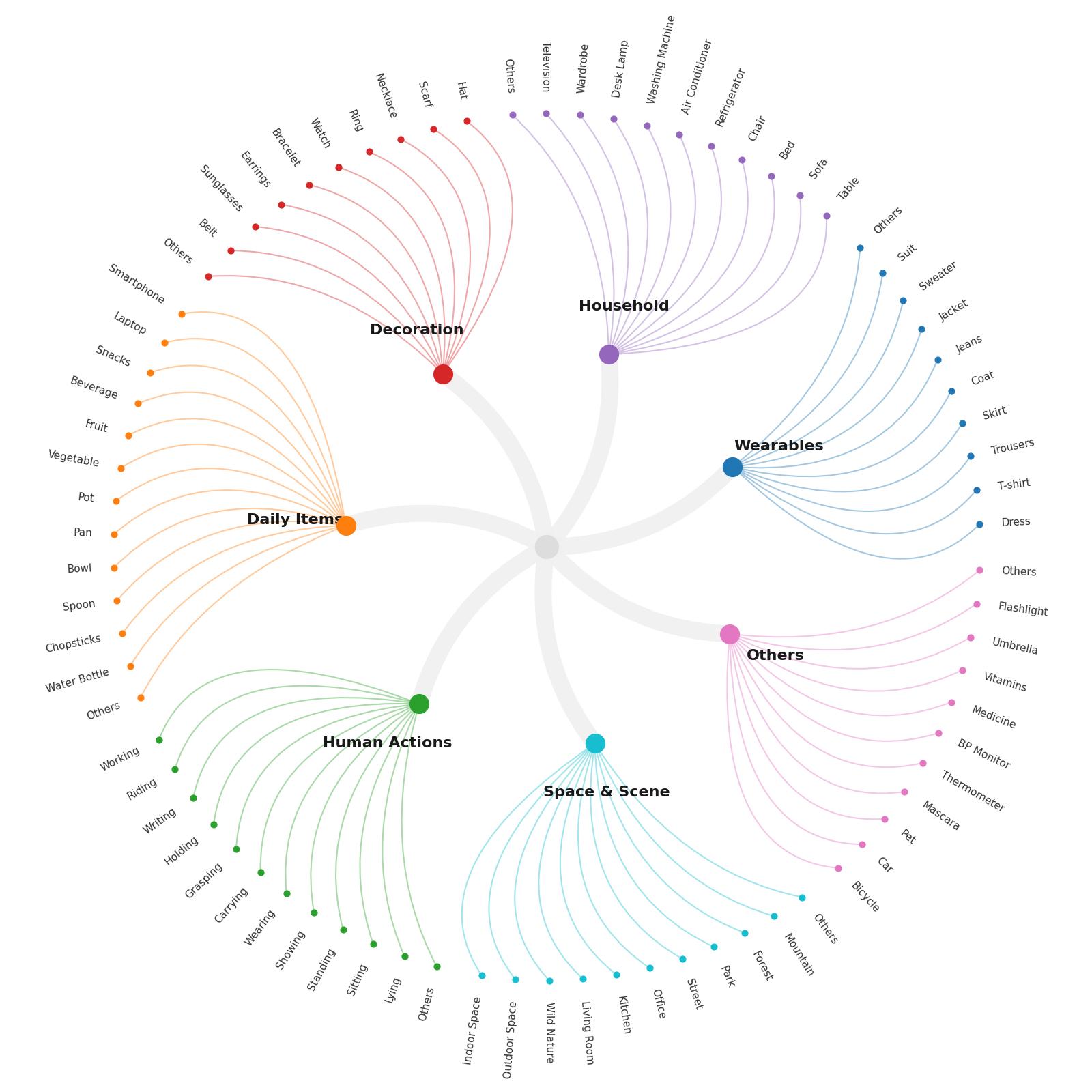}
        \caption{The dataset contains diverse categories.}
        \label{fig:ca}
    \end{minipage}
\end{figure*}
To further illustrate the critical tension between long-form reasoning and visual grounding discussed in the main text, we provide a detailed qualitative breakdown of the reward generation process in Fig.~\ref{fig:model2}. In this watch insertion scenario, baseline MLLM methods such as Direct Scoring, CoT Mean, and Logit Weighted Scoring output artificially inflated scores (ranging from 4.4 to 5). This occurs because the language model falls into a fluent reasoning hallucination, convincingly describing excellent lighting and seamless integration while completely missing a fatal geometric flaw: the watch face is rotated sideways, making it unreadable for the runner.

By contrast, our Evaluation Verification Reward (EVR) explicitly disentangles this process. As shown in Fig.~\ref{fig:model2}, the Evaluator first generates multiple independent short reasoning hypotheses. While several hypotheses contain hallucinated praises and invalid perfect scores, the Verifier systematically cross-checks each specific claim against concrete visual evidence. It successfully identifies that the screen orientation is physically implausible. By rejecting the ungrounded evaluations and retaining only the factually correct critique, EVR outputs a reliable, visually grounded final score of 3. This perfectly demonstrates our core claim: explicitly verifying factual assertions prevents the reward model from being blinded by text-induced bias.

This qualitative behavior perfectly explains the macroscopic statistical properties observed across our preference dataset. As detailed in Table.~\ref{tab:reward_stats} and visualized in Fig.~\ref{fig:reward_dist}, baseline methods suffer from severe context bias and score inflation. Direct Scoring yields an overconfident mean of 0.95 with exceptionally low variance (0.029), essentially acting as a perfect score generator regardless of actual edit quality. Similarly, ensemble and logit-based methods maintain artificially high means around 0.80. In stark contrast, the EVR framework achieves a mean reward of 0.63 with a healthy sample variance of 0.073. Rather than indicating poorer performance, this lower mean, combined with higher variance, proves that EVR successfully deflates hallucinated high scores, offering the highly discriminative and trustworthy reward signal necessary for stable diffusion policy learning.

\subsection{Detailed results of User Study}
\begin{figure}[ht!]
  \centering
  \includegraphics[width=\linewidth]{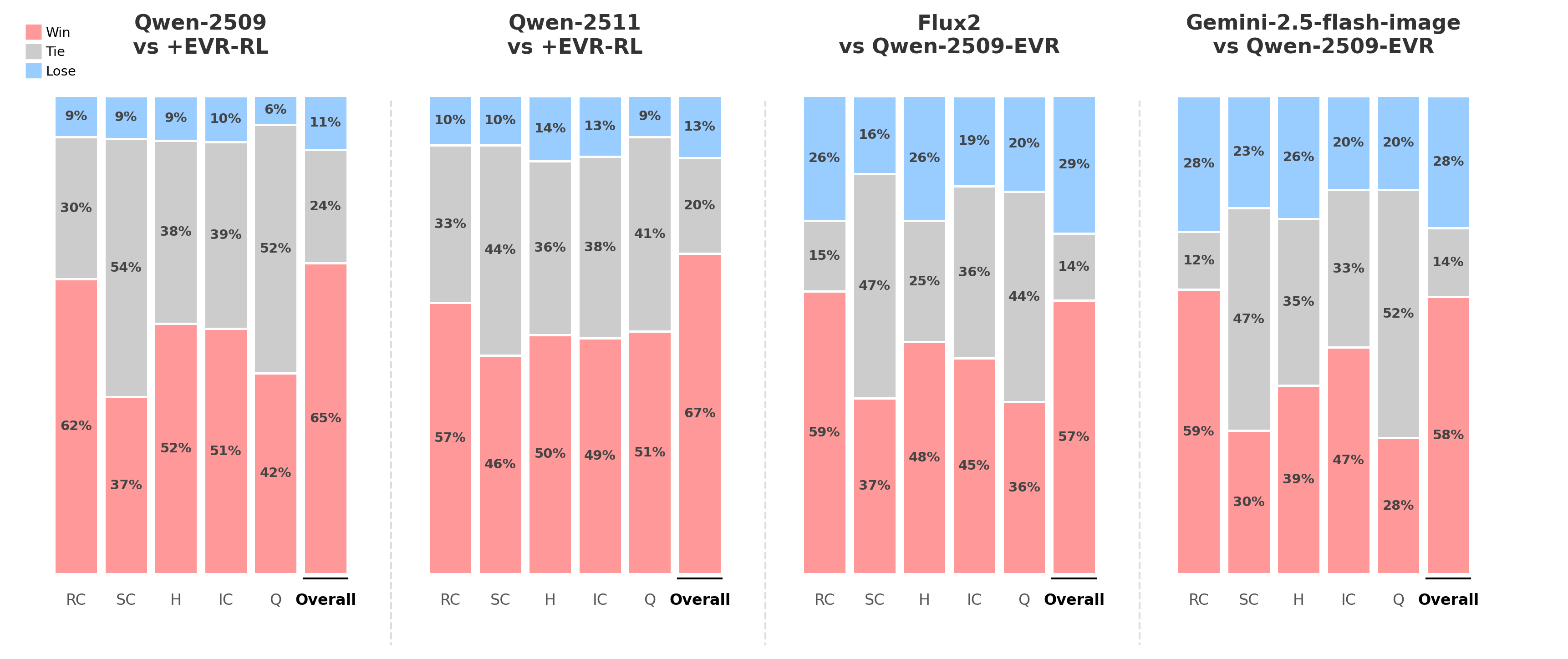}
  \caption{Detailed results of user study}
  \label{fig:user2}
\end{figure}
We conducted a comprehensive user study with random assignment, involving 12 expert evaluators, to compare our EVR policy against base models (Qwen-2509, Qwen-2511) and state-of-the-art systems (Flux2, Gemini-2.5-flash-image). As shown in Fig.~9, our method demonstrates a decisive advantage. Against Qwen-2509 and Qwen-2511, we achieve strict win rates of 65.0\% and 67.0\%, respectively, securing a win-to-loss ratio of nearly 6 to 1 against the former. Furthermore, our policy maintains strict superiority over powerful generalist models, winning 57.2\% against Flux2 and 58.0\% against Gemini.

The detailed dimension breakdown further validates our multi dimensional reward design, showing particular strength in Reference Consistency (RC) and Harmony (H). Compared to Qwen-2509, our model achieves win rates of 62\% in RC and 52\% in Harmony. Even against the highly capable Gemini model, our method sustains a 59\% win rate in RC, proving that our verification mechanism effectively enforces complex relational constraints that baseline models fail to capture.

\subsection{Cross-Model and Strategy Analysis.}
To further validate EVR's robustness, we compare six configurations across model scales (8B/32B/3.5-Plus) and reasoning strategies (verification vs. CoT-only). As shown in Table~\ref{tab:cross_model}, EVR consistently outperforms unverified baselines on \textit{Consistency} and \textit{Harmony}, the two dimensions most prone to text-induced hallucinations. Notably, Qwen3-VL-8B-Instruct, equipped with EVR, matches or exceeds 32B/3.5-Plus baselines without verification on key dimensions, confirming that the verification pipeline effectively compensates for limited model capacity and delivers reliable rewards regardless of backbone scale.

\begin{table}[t]
\centering
\caption{Quantitative comparison across model scales and strategies. Metrics are calculated by Qwen3.5-plus API}
\label{tab:cross_model}
\resizebox{\linewidth}{!}{
\begin{tabular}{lcccccc}
\toprule
\textbf{Config} & \textbf{Consistency} & \textbf{Scene} & \textbf{Harmony} & \textbf{Instruction} & \textbf{Quality} & \textbf{Reward} \\
\midrule
Qwen3-VL-8B-Instruct(w/o verify) & 4.11 & 4.57 & 2.52 & 3.93& 4.18 & 0.66 \\
Qwen3-VL-8B-Instruct(EVR) & 4.36 & 4.51 & 2.63 & 3.98 & 4.15 & 0.68 \\
\midrule
Qwen3-VL-32B-Instruct(w/o verify)  & 4.16 & 4.61 & 2.58 & 3.99  & 4.17 & 0.67\\
Qwen3-VL-32B-Instruct(EVR)& 4.27 & 4.41 & 2.57 & 4.03  & 4.18 & 0.68\\
\midrule
Qwen3.5-plus(w/o verify) & 4.29 & 4.53 & 2.59 & 4.01 & 4.18 & 0.68 \\
Qwen3.5-plus(EVR)  & 4.43 & 4.67 & 2.73 & 4.01 & 4.18 & 0.70 \\
\bottomrule
\end{tabular}
}
\end{table}

\subsection{Discussion on Computational Efficiency}
\label{sec:appendix_cost}
\begin{wrapfigure}{r}{0.36\linewidth}
    \centering
    \vspace{-1.0em}
    \includegraphics[width=0.96\linewidth]{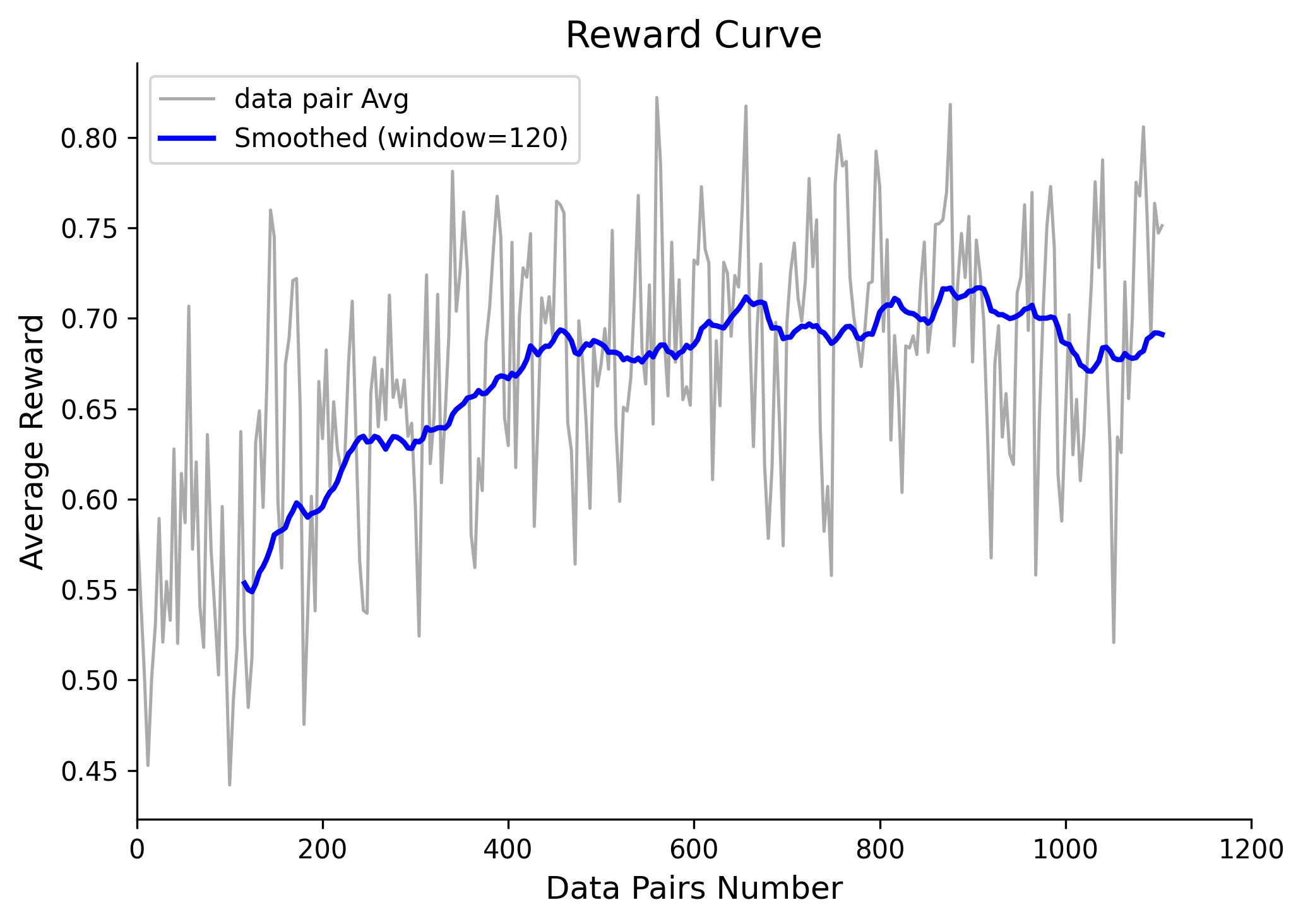}
    \vspace{-1.0em}
    \caption{Training Curve.}
    \label{fig:train_curve}
    \vspace{-0.6em}
\end{wrapfigure}
Although the Evaluation-Verification Reward (EVR) mechanism involves a two-stage process (generating multiple hypotheses and verifying them), our implementation ensures minimal overhead through 2 key optimizations.

\paragraph{Data Efficiency.}
As shown in Fig.~\ref{fig:train_curve}, our system converges with only 1,000 training samples, significantly reducing the total number of reward queries required over the entire training trajectory.

\paragraph{Asynchronous Inference Pipeline.}
We deploy the MLLM Evaluator and Verifier using \textbf{vLLM}~\citep{kwon2023efficient} for local serving. By leveraging \textbf{asynchronous request handling}, the training loop dispatches reward computation tasks without blocking the GPU used for diffusion model updates. This design achieves real-time reward generation with negligible waiting time, effectively overlapping inference with gradient computation.

Consequently, the marginal increase in per-step wall-clock time is negligible. The substantial gains in editing consistency and harmony achieved by EVR thus come at a minimal practical cost.
\subsection{\texorpdfstring{Failure Case when $N\geqslant 5$}{Failure Case when N >= 5}}
\begin{figure}[ht!]
  \centering
  \includegraphics[width=\linewidth]{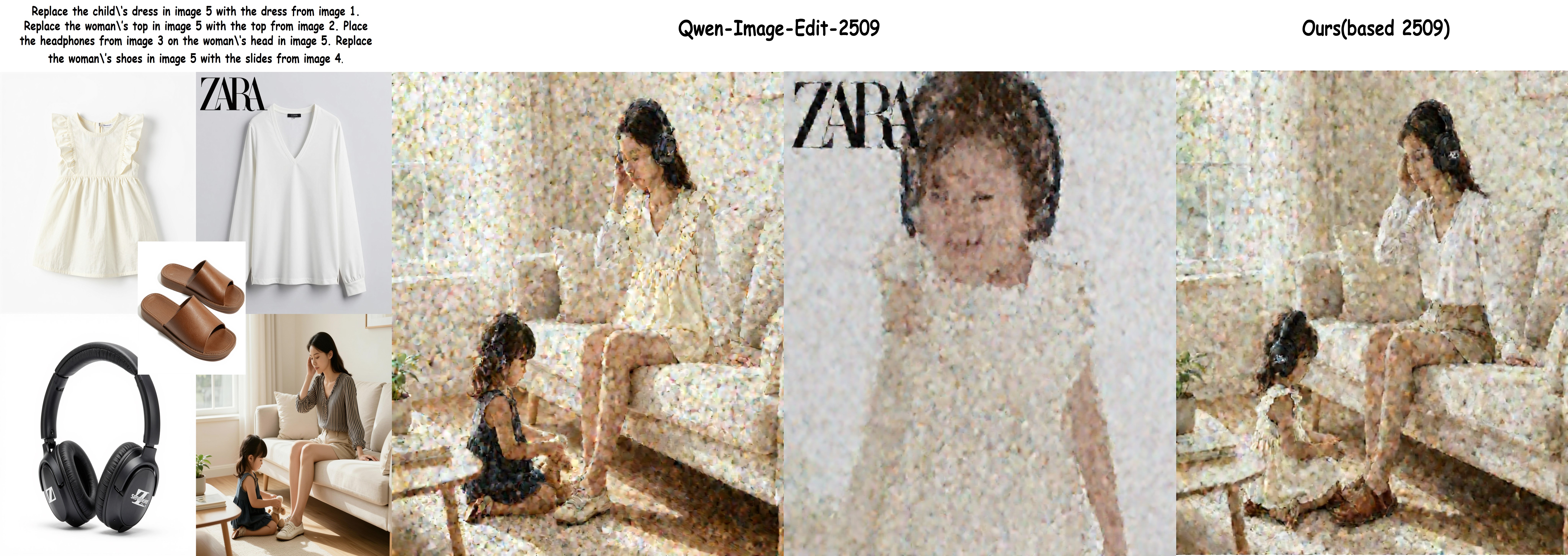}
  \caption{The performance of Qwen-Image-Edit-2509 and our method when $N=5$. The first two images from the base model show noticeable noise. While our result also retains some noise, detailed observation confirms the successful placement of reference objects: the children's outfit on the child, the top on the woman, and the addition of slippers and headphones. This indicates that while our method does not resolve the base model's denoising issues in complex scenes, it markedly improves semantic consistency and visual harmony.}
  \label{fig:failure}
\end{figure}
To further explore the boundaries of our framework, we investigate extreme scenarios where the number of reference images is exceptionally large ($N \ge 5$). As discussed in our Limitations section, while our approach significantly optimizes instruction alignment and relational constraints, it remains fundamentally bounded by the generative prior of the underlying base model.

When tasked with integrating five or more distinct references, the base models begin to exceed their fundamental operational capacity, as shown in Fig.~\ref{fig:failure}. Under these extreme conditions, we observe severe degradation in basic image quality, frequently manifesting as feature entanglement across different objects and incomplete denoising processes. These fundamental collapses indicate that the sheer complexity of the task overwhelms the raw synthesis capabilities of the diffusion architecture. 

Interestingly, despite our fine-tuned method exhibiting similar noise artifacts as the base model in these extreme cases, a closer inspection reveals that our framework successfully places the reference objects into their appropriate spatial and semantic contexts. Specifically, the children's clothing is correctly fitted onto the child, the top is appropriately worn by the woman, and accessories such as slippers and headphones are successfully integrated. This demonstrates that while our method cannot compensate for the base model's failure to denoise highly complex scenes, it still significantly elevates semantic consistency and visual harmony.

Consequently, while our Evaluation-Verification Reward consistently improves the overall multi-image editing results within the capability limits of the base model, it cannot completely rescue the generation in terms of low-level image fidelity when the foundational model suffers from absolute prior collapse. This observation directly supports our limitation analysis: our EVR mechanism provides highly reliable optimization gradients for relational alignment, but fully solving extreme-scale multi-reference integration will ultimately require base models with stronger native generative capacities.

\subsection{More Results}

To further demonstrate the robustness and versatility of our proposed EVR framework, we present additional qualitative results in this section, as shown in Fig.~\ref{fig:app11}, \ref{fig:app21}, \ref{fig:ood2}, and \ref{fig:ood3}. These supplementary examples highlight our model's capability to consistently maintain reference identity and visual harmony across a diverse range of complex editing instructions.

%% file: sections/appendix/c_release_checklist.tex
\section{Prompt Templates}

To facilitate full reproducibility and provide complete transparency regarding our Evaluation Verification Reward framework, we present the exact system prompts utilized in our experiments. As discussed in the Methodology section, the Evaluator prompt is specifically engineered to elicit structured reasoning across five distinct visual dimensions. Conversely, the Verifier prompt strictly enforces visual grounding by instructing the multimodal model to accept or reject these generated claims based solely on concrete pixel evidence.

\subsection{Evaluator Prompt}
The following prompt is used to instruct the base language model to act as the Evaluator, generating multiple independent rationales and preliminary scores.

\begin{promptbox}[title=Reference Consistency]
Your role is to evaluate whether a specific object present in the reference image is correctly retained in the edited image. The task is not to assess global similarity, but to verify that the edited image contains the same object from Reference Image, matching in identity and essential visual characteristics.

The edited image is generted by {prompt}, It involve a object insertion or replacement.
Please only focus on the items that coexist in reference image and the edited image, ignoring other elements in the edited image.

Evaluating consistency is not pixel level, but rather from aspects such as material, shape, arrangement, color, lighting, composition, etc

1. No Match: The object from Reference Image is completely absent in Edited Image, or a different object (different category or identity) appears in its place.

2. Weak Match: An object of the same general category is present (e.g., “a chair” instead of “the red office chair”), but key identifying features (color, shape, style, brand, or distinctive markings) are missing or altered beyond recognition.

3. Partial Match: The correct object is likely intended, but critical attributes are inconsistent (e.g., wrong color, material changed, major design elements missing), making it visually distinct from the reference. Excessive structure or ghosting appears.

4. Match: The object in Edited Image closely matches the one in Reference Image in type, form, color, texture, and distinctive details; minor differences due to lighting, viewpoint, partial occlusion, or editing artifacts are acceptable.

5. Exact Match: The object in Edited is visually faithful to the one in Reference Image across all defining characteristics—same model, condition, color, pose, and unique identifiers—with only natural variations expected from image editing (e.g., perspective shift, brightness adjustment).

Response Format (Please specify the reason for the deduction within the <Reason> tag. Respond to the score directly within the <Score> tag.):

<Reason>...</Reason><Score>1-5</Score>
\end{promptbox}

\begin{promptbox}[title=Scene Consistency]
You are evaluating scene-level consistency between a scene image and an edited image. The edit involves only a localized change (e.g., adding or replacing an object like a watch, bag, or clothing).

The Edited image is generated by {prompt}, It involve a object insertion or replacement. Ignore the edited object itself. 
Do not judge based on changed meaning or style due to the edited object.
Instead, assess whether the rest of the scene—including environment, lighting, perspective, and any human subjects (pose, anatomy, orientation)—remains physically coherent and visually unchanged.

Pay special attention to:
Human pose and body integrity (no distortions or unnatural limbs),
Background stability (no warping, ghosting, or foreign textures),
Consistent lighting, shadows, and camera perspective,
No leakage of background or style from the source of the inserted object.

Scoring Criteria (focus: unchanged scene integrity):

1. Inconsistent: The non-edited parts of the scene are fundamentally altered—e.g., indoor/outdoor switch, distorted room geometry, warped background, or human anatomy broken; clear contamination from the source object’s context.

2. Mostly Inconsistent: Scene type is recognizable, but major artifacts in unchanged areas: severe pose distortion, mismatched global lighting, perspective shift, or visible background bleed affecting realism.

3. Partially Consistent: Core structure is preserved, but subtle issues in non-edited regions—e.g., slight pose drift, inconsistent ambient occlusion, minor texture warping, or soft lighting mismatches—reduce plausibility.

4. Mostly Consistent: Unchanged content (scene + subject) remains visually stable; lighting, perspective, and materials align well. Only negligible imperfections (e.g., tiny shadow edge discrepancy) that don’t break immersion.

5. Fully Consistent: The entire non-edited portion is indistinguishable from the original—perfect preservation of pose, environment, lighting, and style. No trace of external contamination or structural alteration.

Response Format (Please specify the reason for the deduction within the <Reason> tag. Respond to the score directly within the <Score> tag.):

<Reason>...</Reason><Score>1-5</Score>
\end{promptbox}

\begin{promptbox}[title=Harmony]
H:  Your role is to evaluate the harmonization quality of a reference object (from Image 1) after it has been inserted or edited. The assessment focuses primarily on the physical and common-sense plausibility of the object within its new environment or the interaction mode between people and objects.

Pay more attention to the wearing mode (such as the screen orientation, the position), grip posture, clipping and violation of gravity.

1. Poor Harmony: The object appears obviously pasted in—sharp cut-out edges, physically impossible positions or poses, implausible scale/proportions for the setting, or semantically absurd placements. collage effect.

2. Fair Harmony: While the object could belong to the plausible category for the scene, there are noticeable issues related to physical rules or common sense, such as objects intersecting unnaturally (clipping issues), or unrealistic interactions between objects. These issues significantly detract from realism despite attempts at blending.

3. Ordinary Harmony: Objects are placed generally conform to the physical logic of the scene. However, it may violate the patterns of daily use, such as slight clipping artifacts or unrealistic object interactions (such as inappropriate wearing patterns, unreasonable orientations, etc., positions that do not conform to daily use), but these will not seriously affect the overall credibility.

4. Good Harmony: The placement and interaction of the object closely follow physical laws and common usage patterns, with only minute flaws that might escape notice by most viewers. Aspects like lighting direction, shadow softness, and color grading are also well-matched. Only expert scrutiny might reveal editing traces.

5. Excellent Harmony: The inserted object is indistinguishable from native scene content, perfectly aligned with physical rules, daily use patterns, and visual consistency. There are no visual cues suggesting compositing—it reads as a single, authentic photograph.

Response Format (Please specify the reason for the deduction within the <Reason> tag. Respond to the score directly within the <Score> tag.):

<Reason>...</Reason><Score>1-5</Score>
\end{promptbox}

\begin{promptbox}[title=Instruction Consistency]
IC:  Your role is to evaluate how faithfully and effectively an editing instruction has been executed when inserting or modifying reference images. The primary focus is on instruction adherence: whether the requested change was carried out accurately.

Image 1: Reference Image 1
Image 2: Reference Image 2
Image 3: Edited Image, your subject to evaluate.
Instruction:\{prompt\}

Please first identify which are reference images and which are edited images, analyze the content presented by the input images separately, and then determine whether the instructions have been followed.

1. Bad Execution: The editing instruction is largely ignored or misinterpreted. The wrong object is inserted, placed in an irrelevant location, or depicted in a way that contradicts the directive (e.g., instructed to “place a coffee cup on the table” but a teapot appears on the floor). Severe violations of physics or context further undermine fidelity.

2. Poor Execution: The general intent of the instruction is recognizable, but key details are inaccurate or incomplete—wrong orientation, incorrect scale, missing required interactions (e.g., cup floating above table instead of resting on it), or partial fulfillment (only part of a multi-step instruction is followed).

3. Fair Execution: The instruction is mostly followed correctly: the right object appears in the specified location with appropriate orientation and basic interaction. Minor deviations exist—such as slight misalignment, weak contact with supporting surfaces, or subtle inconsistencies in usage context—but the core request is fulfilled.

4. Good Execution: The instruction is executed with high precision. Object identity, position, pose, and interaction align closely with the directive. Physical contact, support, and spatial relationships appear natural. Visual blending (lighting, shadows, color) supports realism, though tiny imperfections may remain upon close inspection.

5. Excellent Execution: The instruction is implemented flawlessly. Every aspect—object type, placement, orientation, physical interaction (e.g., proper support, no clipping), and semantic context—matches the directive exactly. The result looks like a native part of the original photograph, with no detectable errors in logic, physics, or execution.

Response Format (Please specify the reason for the deduction within the <Reason> tag. Respond to the score directly within the <Score> tag.):

<Reason>...</Reason><Score>1-5</Score>
\end{promptbox}

\begin{promptbox}[title=Visual Quality]
Your role is to evaluate the aesthetic quality score of given images.

1. Bad: Extremely blurry, underexposed with significant noise, indiscernible subjects, collapsed subjects, and chaotic composition.

2. Poor: Noticeable blur, poor lighting, washed-out colors, and awkward composition with cut-off subjects, incomplete object .

3. Fair: In focus with adequate lighting, dull colors, decent composition but lacks creativity.

4. Good: Sharp, good exposure, vibrant colors, thoughtful composition with a clear focal point.

5. Excellent: Exceptional clarity, perfect exposure, rich colors, masterful composition with emotional impact.

Response Format (Please specify the reason for the deduction within the <Reason> tag. Respond to the score directly within the <Score> tag.):

<Reason>...</Reason><Score>1-5</Score>
\end{promptbox}

\subsection{Verifier Prompt}
The following prompt configures the Verifier to cross check the Evaluator outputs against the visual inputs, ensuring that only factually grounded claims contribute to the final reward signal.

\begin{promptbox}[title=Verifier]
Verifier:  You are an Image Evaluation Validator responsible for verifying model-generated assessments of edited images. Your task is to review these evaluations based on specific criteria provided in the original evaluation task.

The original evaluation task involves: \{task\}

You will be given:
- the reference and edited images
- A list of model-generated evaluations, each in the format: <Reason>[...]</Reason><Score>X</Score>.
\{prompt\}

The following is several evaluations:
\{evaluations\}

Your validation must follow this three-step process:

1. **Aggregate Claims**: Extract every factual claim, observation, or alleged flaw mentioned across all <Reason> sections.

2. **Verify Against Evidence**: According to the edited image, determine which of these claims are actually true, false, or unsupported. 
The deduction statements regarding wearing patterns and physical laws in each evaluation may not be groundless. Please treat these evaluations with caution and carefully examine them.

3. **Evaluate Each Assessment**: For each original evaluation, check:

- Whether its reasoning relies only on verified facts,

- Whether it correctly applies the rubric definitions,

- Whether its final score logically follows from accurate observations,

- Whether it fails to mention a critical flaw that is both present and relevant to the rubric.

Only if all conditions are met should the evaluation be marked as valid (1). Otherwise, mark it as invalid (0).

Provide your full reasoning in the <Thought> tag. Then output a list of 0s and 1s in the exact order of the input evaluations inside the <Answer> tag. Do not include any extra text, formatting, or punctuation beyond what is required.

Output Format:

<Thought>[Your step-by-step verification and analysis]</Thought>

<Answer>[1,0,1,...]</Answer>
\end{promptbox}

%% file: sections/appendix/d_additional_qualitative_results.tex
\section{Additional Qualitative Results}

\begin{figure}[p]
    \centering
    \includegraphics[width=\linewidth]{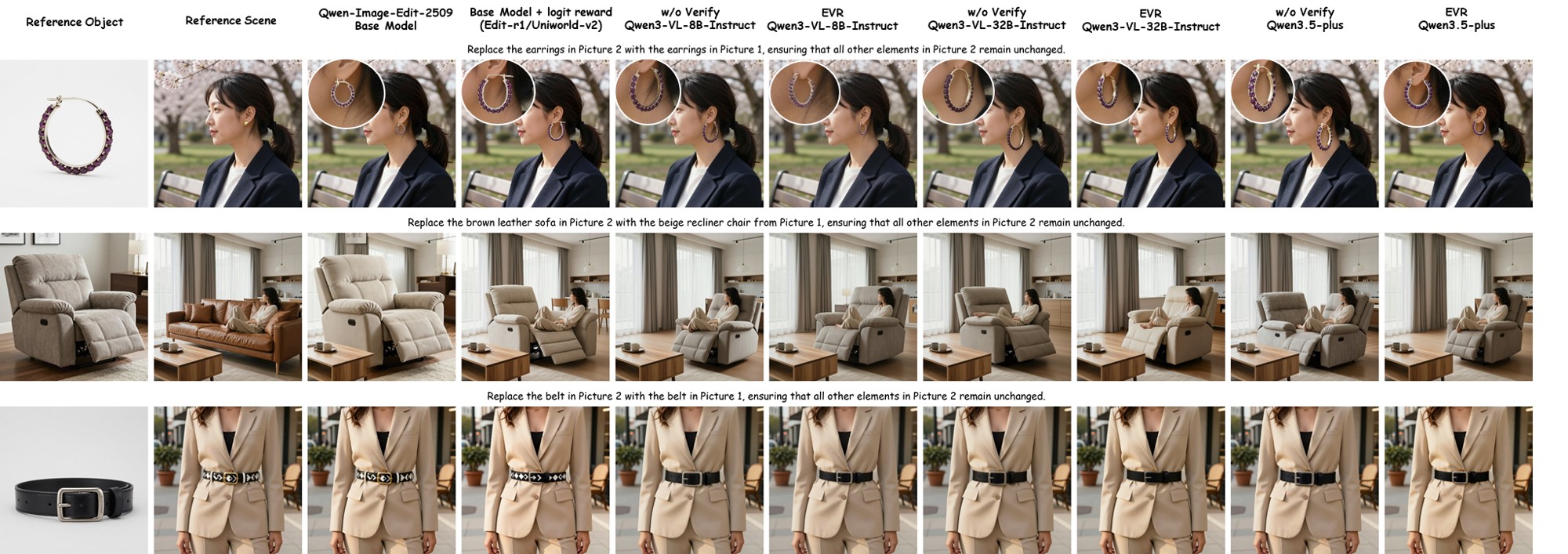}
    \caption{Qualitative comparison with other MLLM-based methods and qualitative comparison of model robustness. Compared to Edit-R1, our method is substantially better. Compared to the setting without verifier, our method exhibits stable harmony and consistency across models of different scales.}
    \label{fig:cross}

    \vspace{1.2em}

    \begin{minipage}{0.5\linewidth}
        \centering
        \includegraphics[width=\linewidth]{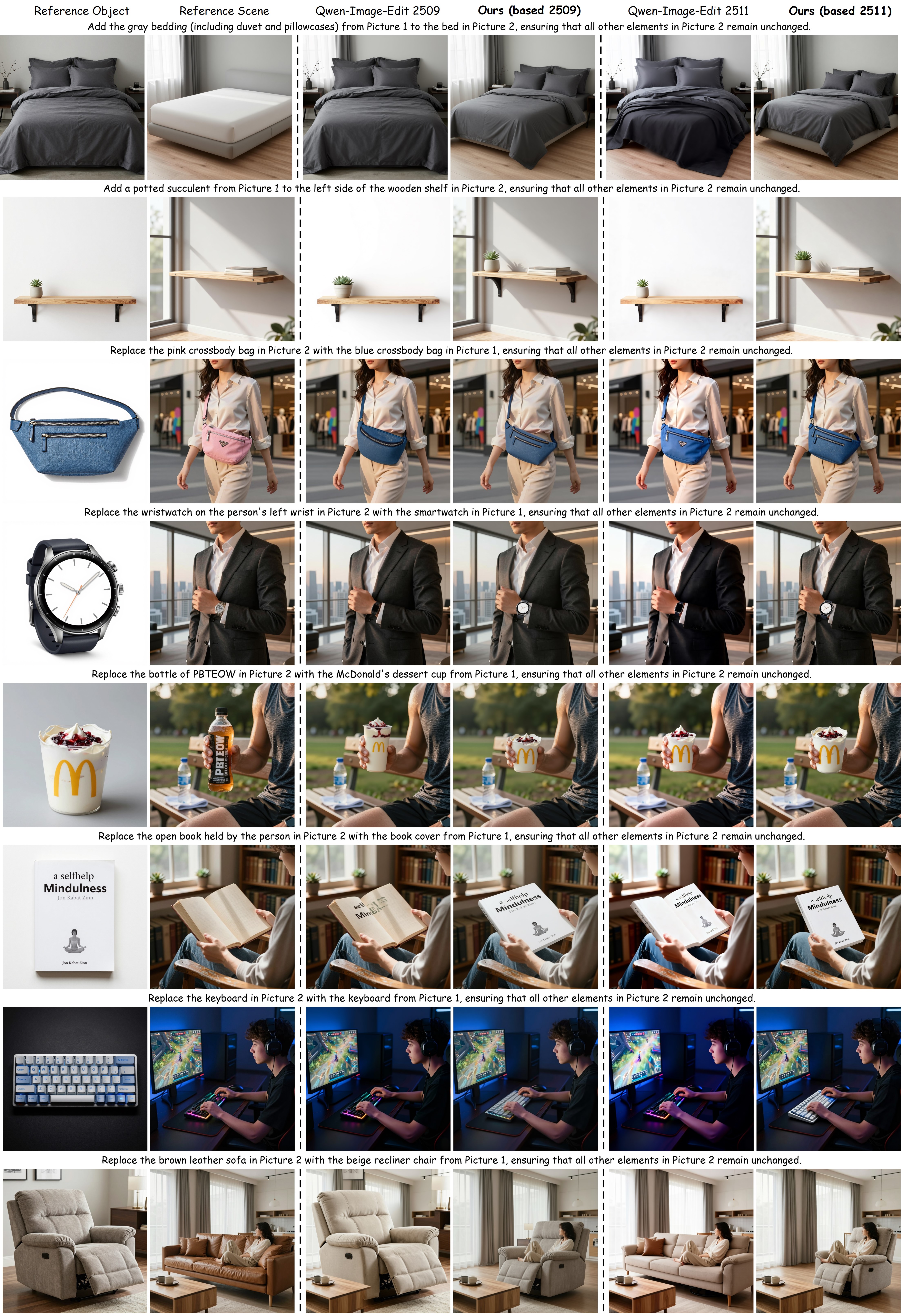}
        \caption{Additional qualitative results.}
        \label{fig:app11}
    \end{minipage}
    \hfill
    \begin{minipage}{0.48\linewidth}
        \centering
        \includegraphics[width=\linewidth]{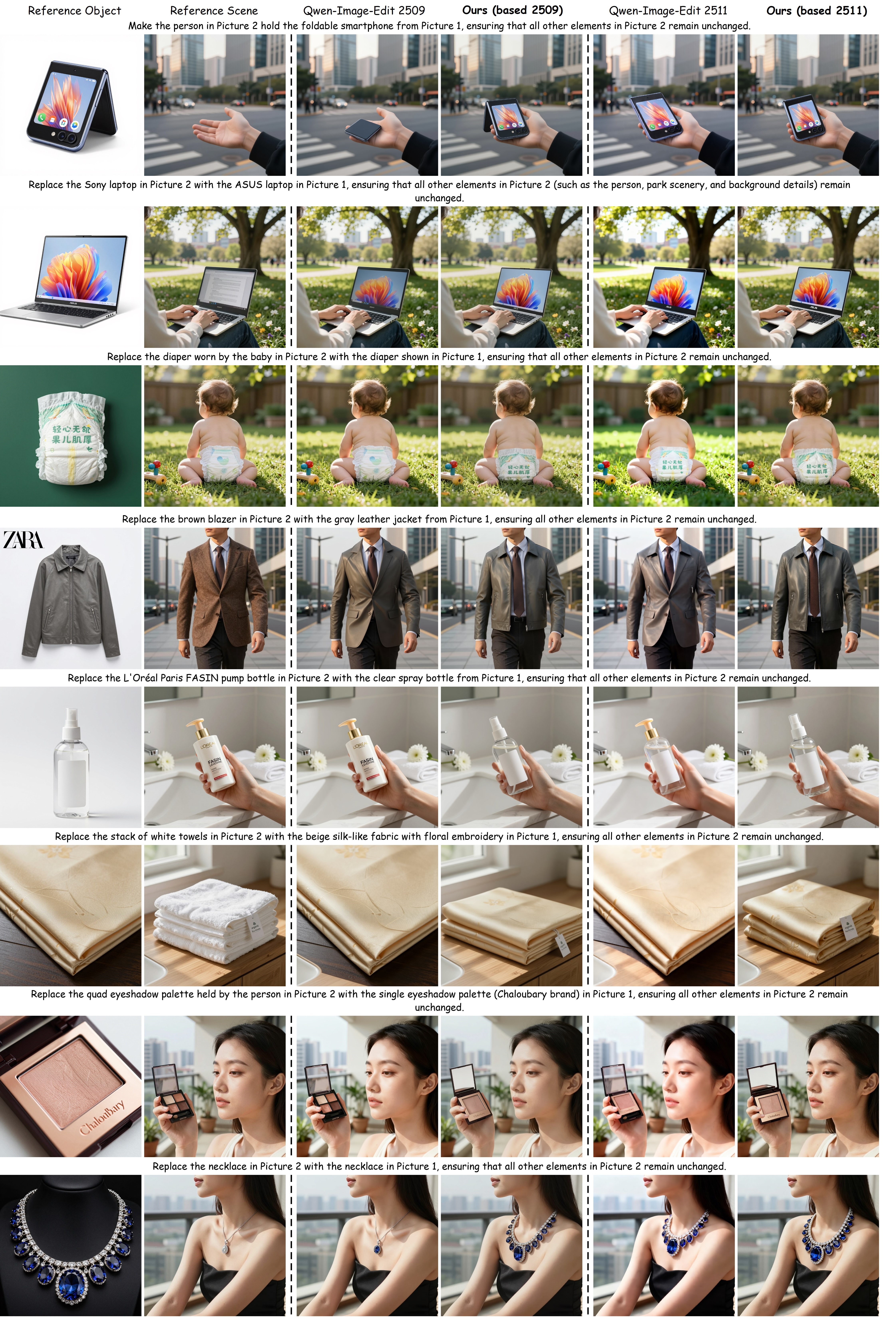}
        \caption{Additional qualitative results.}
        \label{fig:app21}
    \end{minipage}
\end{figure}

\begin{figure}[p]
    \centering
    \includegraphics[width=\linewidth]{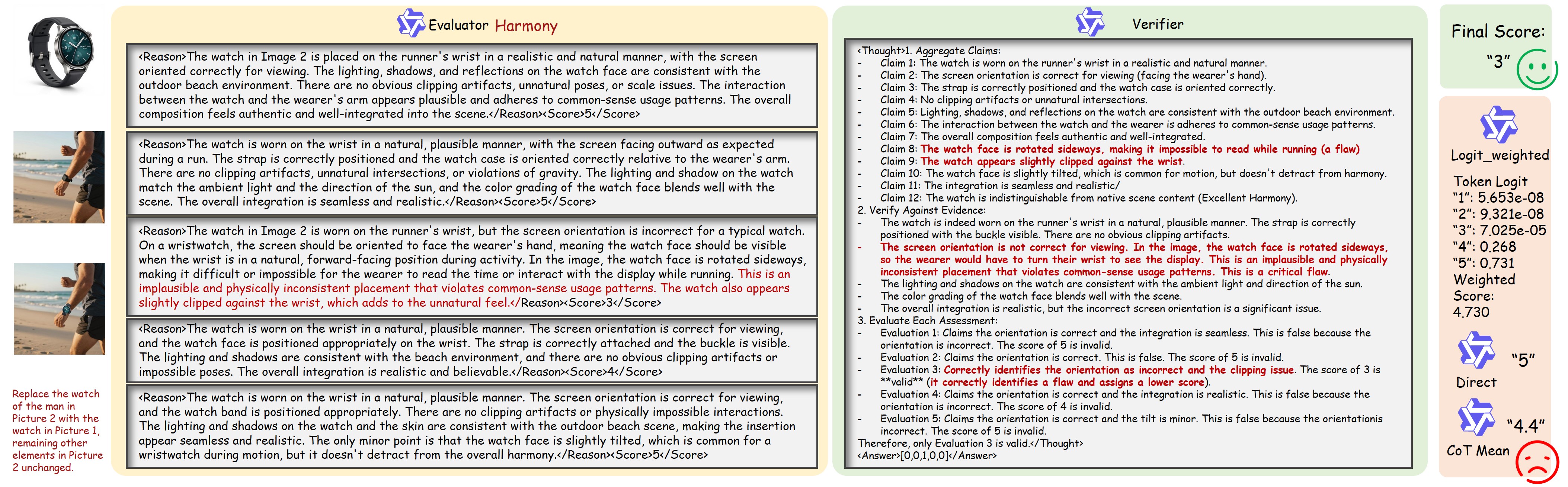}
    \caption{Comparison between EVR and other MLLM-based reward models. The first row shows the input images and instructions, while the second row displays the outputs and their corresponding evaluation by different methods. It illustrates how EVR suppresses hallucinations compared to other reward formulations.}
    \label{fig:model2}

    \vspace{1.2em}

    \begin{minipage}{0.48\linewidth}
        \centering
        \includegraphics[width=\linewidth]{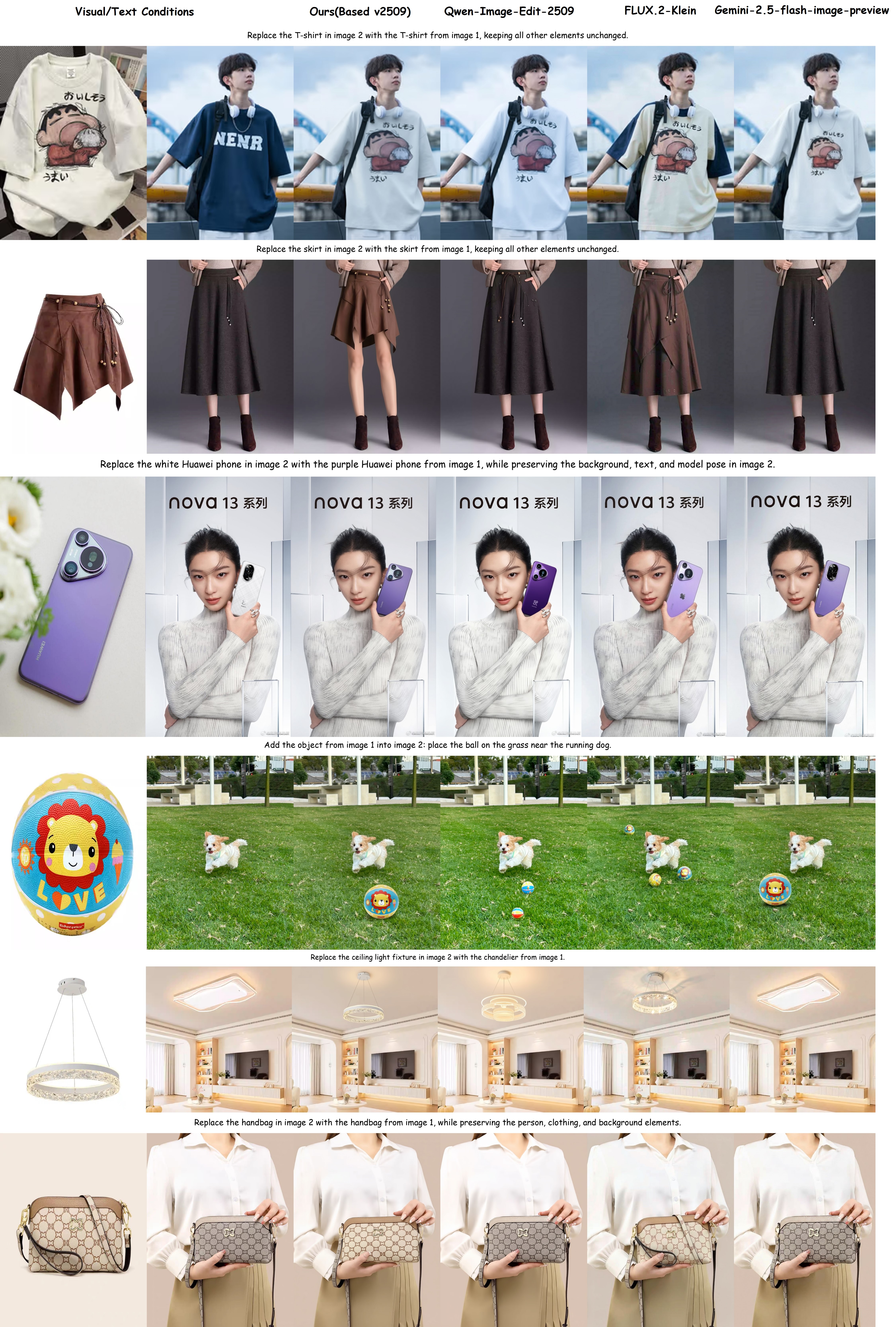}
        \caption{Additional results on the OOD dataset.}
        \label{fig:ood2}
    \end{minipage}
    \hfill
    \begin{minipage}{0.5\linewidth}
        \centering
        \includegraphics[width=\linewidth]{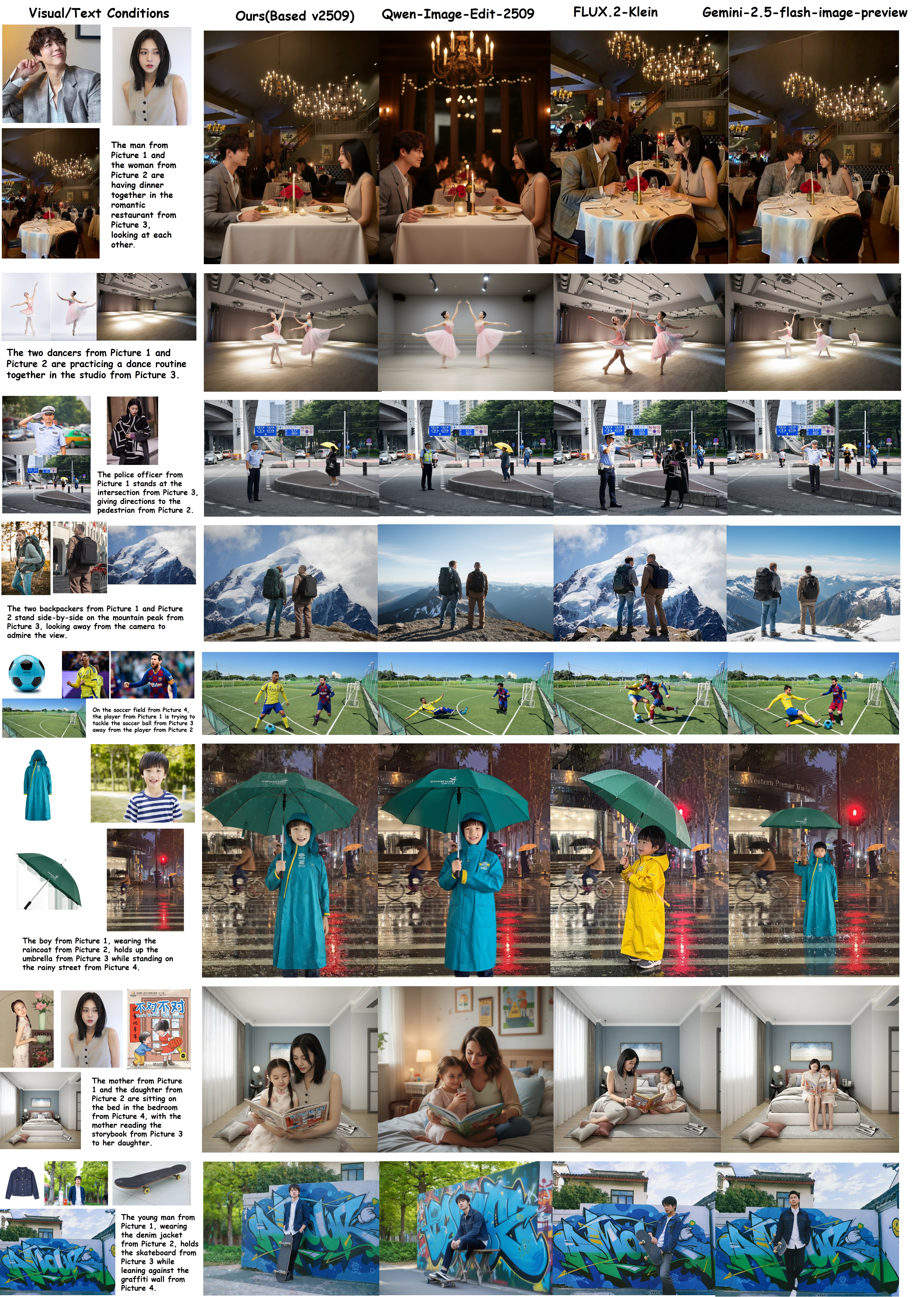}
        \caption{Additional results on the $N>2$ dataset.}
        \label{fig:ood3}
    \end{minipage}
\end{figure}

This appendix provides additional qualitative evidence for the main claims in the paper. Fig.~\ref{fig:cross} compares EVR with alternative MLLM-based reward strategies and verifies its robustness across evaluator scales. Fig.~\ref{fig:model2} further illustrates how EVR suppresses hallucinated reward signals through visual verification. The remaining examples extend the qualitative analysis to additional in-distribution, OOD, and $N>2$ editing cases.

